\documentclass{article}
\pdfoutput=1
\usepackage{microtype}
\usepackage{graphicx}
\usepackage{subfig}
\usepackage{booktabs} 
\usepackage{amsmath,amssymb,amsthm}
\usepackage{pgfplots}
\usepackage{verbatim}
\usepackage{enumitem}

\usepackage[mode=buildnew]{standalone}

\pgfplotsset{compat=newest}
\DeclareUnicodeCharacter{2212}{−}
\usepgfplotslibrary{groupplots,dateplot}
\usetikzlibrary{patterns,shapes.arrows}
\usepgfplotslibrary{fillbetween}

\graphicspath{{.}{./figs/}}

\colorlet{myblue}{blue!50}
\newcommand{\lightercolor}[3]{
    \colorlet{#3}{#1!#2!white}
}

\newcommand{\darkercolor}[3]{
    \colorlet{#3}{#1!#2!black}
}

\lightercolor{myblue}{50}{lightblue}
\darkercolor{myblue}{50}{darkblue}
\newlength{\mylinewidth}
\newcommand{\tr}{{\rm tr}}
\newcommand{\T}{{\sf T}}
\renewcommand{\H}{{\sf H}}

\newtheorem{assumption}{Assumption}
\newtheorem{theorem}{Theorem}
\newtheorem{remark}{Remark}

\usepackage{hyperref}

\usepackage[accepted]{icml2021}

\icmltitlerunning{Two-way kernel matrix puncturing}

\begin{document}

\twocolumn[
\icmltitle{\texorpdfstring{Two-way kernel matrix puncturing:\\ towards resource-efficient PCA and spectral clustering}{Two-way kernel matrix puncturing: towards resource-efficient PCA and spectral clustering}}



\icmlsetsymbol{equal}{*}

\begin{icmlauthorlist}
\icmlauthor{Romain Couillet}{gipsa,cs}
\icmlauthor{Florent Chatelain}{gipsa}
\icmlauthor{Nicolas Le Bihan}{gipsa}
\end{icmlauthorlist}

\icmlaffiliation{gipsa}{GIPSA-lab, CNRS, Grenoble-INP, University Grenoble-Alps}
\icmlaffiliation{cs}{CentraleSupélec, University Paris Saclay}

\icmlcorrespondingauthor{Romain Couillet}{romain.couillet@gipsa-lab.grenoble-inp.fr}
\icmlcorrespondingauthor{Florent Chatelain}{florent.chatelain@gipsa-lab.grenoble-inp.fr}
\icmlcorrespondingauthor{Nicolas Le Bihan}{nicolas.lebihan@gipsa-lab.grenoble-inp.fr}

\icmlkeywords{PCA, spectral clustering, random matrix theory, performance-complexity tradeoff}

\vskip 0.3in
]



\printAffiliationsAndNotice{\icmlEqualContribution} 

\begin{abstract}
The article introduces an elementary cost and storage reduction method for spectral clustering and principal component analysis. The method consists in randomly ``puncturing'' both the data matrix $X\in\mathbb{C}^{p\times n}$ (or $\mathbb{R}^{p\times n}$) and its corresponding kernel (Gram) matrix $K$ through Bernoulli masks: $S\in\{0,1\}^{p\times n}$  for $X$ and $B\in\{0,1\}^{n\times n}$ for $K$. The resulting ``two-way punctured'' kernel is thus given by $K=\frac1p[(X\odot S)^\H (X\odot S)]\odot B$.
We demonstrate that, for $X$ composed of independent columns drawn from a Gaussian mixture model, as $n,p\to\infty$ with $p/n\to c_0\in(0,\infty)$, the spectral behavior of $K$ -- its limiting eigenvalue distribution, as well as its isolated eigenvalues and eigenvectors -- is fully tractable and exhibits a series of counter-intuitive phenomena. We notably prove, and empirically confirm on various real image databases, that it is possible to drastically puncture the data, thereby providing possibly huge computational and storage gains, for a virtually constant (clustering or PCA) performance. This preliminary study opens as such the path towards rethinking, from a large dimensional standpoint, computational and storage costs in elementary machine learning models.
\end{abstract}

\section{Introduction}
\label{sec:intro}

The ever-increasing tremendous amounts of data that machine learning algorithms now need to face start to tip the scale towards a major computational and storage resource bottleneck. In such fields as astrophysics with the recent SKA radiotelescope or Internet data mining, the collected data are simply too large to be stored and must therefore be processed in real-time before being discarded altogether. In parallel, even if those data could be stored, algorithm complexities beyond linear can in general not be afforded. This is already a problem for as elementary methods as principal component analysis (PCA) or spectral clustering -- both related to Gram matrix eigenvector retrieval.

\medskip

Evidently, numerous works have proposed various directions of cost-efficient methods for PCA and spectral clustering. For instance, the line of works \cite{johnstone2009sparse,cai2013sparse,deshpande2014information} provides a series of \emph{sparse PCA} methods by assuming that the principal components are sparse: the main gain arises from automatically selecting the reduced set of covariates having largest amplitude. More recently, inspired by statistical physics, \cite{zhong2020empirical} proposes an \emph{empirical Bayes} version of PCA, by setting a (non-Gaussian) product measure prior on the principal components: \cite{zhong2020empirical} in particular obtains (in simulations) a thousand-fold reduction in the number of data necessary to maintain equal performance with respect to standard PCA. Yet, the most popular methods to handle large dimensional PCA fall into the realm of \emph{dimensionality reduction} and \emph{random projections} \cite{freund2007learning} which, one way or another, also require prior knowledge on the sought principal components to avoid dramatic performance losses. Similar ideas have been devised for spectral clustering, such as hierarchical clustering \cite{murtagh2012algorithms}.

But these works all exploit strong structural prior on the data (e.g., a prior on principal components) to reduce the effective data dimension, and in general only operate on one dimension -- either the data size or number.

\medskip

As for mitigating storage constraints, clustering can be performed in a streaming manner, as proposed in \cite{keriven2018sketching} by means of a \emph{data sketching} approach. This approach however looses much discriminating power in not effectively ``comparing'' all raw data and thus fails to compete against spectral methods. Stochastic gradient descent in deep neural networks also performs clustering in a non-spectral manner by ``streaming'' in small data batches \cite{bottou1991stochastic}, but these algorithms only converge after multiple epochs, meaning that the data must be stored for later reuse. More conventionally, since the addition of new data induce successive rank-$1$ perturbations of the sample covariance, iterative perturbation methods based on the Sherman-Morrison formula can be exploited \cite{engel2004kernel}, however here again at the cost of full data storage.

\medskip

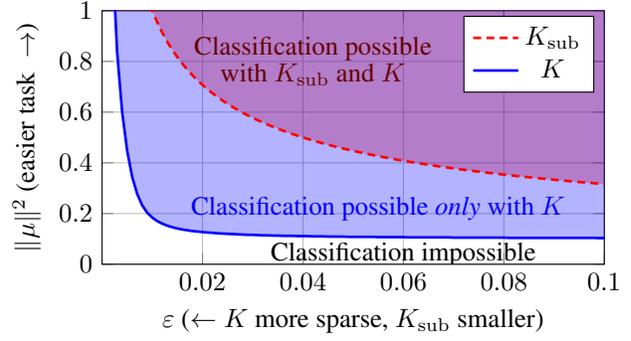
\begin{figure}
\centering
\begin{tikzpicture}
\begin{axis}[
  width=\linewidth,
  height=.6\linewidth,
  xlabel={$\varepsilon$ ($\leftarrow K \textmd{ more sparse, }K_{\rm sub} \textmd{ smaller}$)},
  ylabel={$\|\mu\|^2$ ($\textmd{easier task } \rightarrow$)},
  xmin=0,
  xmax=.1,
  ymin=0,
  ymax=1,
  xtick={.02,.04,.06,.08,.1},
  xticklabels={$0.02$,$0.04$,$0.06$,$0.08$,$0.1$},
  grid=major
]

    \def\C_{.01}
    \addplot[domain=0:.1, samples=100,red,densely dashed,line width=1pt,name path=g]{sqrt(\C_/x)};
    \addlegendentry{$K_{\rm sub}$};
    \addplot[blue,line width=1pt,name path=f] plot coordinates {
	    (0.000,12.000000)(0.001,2.164000)(0.002,1.241000)(0.003,0.836000)(0.004,0.600000)(0.005,0.447000)(0.006,0.344000)(0.007,0.277000)(0.008,0.232000)(0.009,0.203000)(0.010,0.184000)(0.011,0.170000)(0.012,0.160000)(0.013,0.152000)(0.014,0.146000)(0.015,0.141000)(0.016,0.137000)(0.017,0.134000)(0.018,0.131000)(0.019,0.129000)(0.020,0.127000)(0.021,0.125000)(0.022,0.124000)(0.023,0.122000)(0.024,0.121000)(0.025,0.120000)(0.026,0.119000)(0.027,0.118000)(0.028,0.117000)(0.029,0.116000)(0.030,0.116000)(0.031,0.115000)(0.032,0.114000)(0.033,0.114000)(0.034,0.113000)(0.035,0.113000)(0.036,0.112000)(0.037,0.112000)(0.038,0.112000)(0.039,0.111000)(0.040,0.111000)(0.041,0.111000)(0.042,0.110000)(0.043,0.110000)(0.044,0.110000)(0.045,0.109000)(0.046,0.109000)(0.047,0.109000)(0.048,0.109000)(0.049,0.108000)(0.050,0.108000)(0.051,0.108000)(0.052,0.108000)(0.053,0.108000)(0.054,0.108000)(0.055,0.107000)(0.056,0.107000)(0.057,0.107000)(0.058,0.107000)(0.059,0.107000)(0.060,0.107000)(0.061,0.106000)(0.062,0.106000)(0.063,0.106000)(0.064,0.106000)(0.065,0.106000)(0.066,0.106000)(0.067,0.106000)(0.068,0.106000)(0.069,0.106000)(0.070,0.105000)(0.071,0.105000)(0.072,0.105000)(0.073,0.105000)(0.074,0.105000)(0.075,0.105000)(0.076,0.105000)(0.077,0.105000)(0.078,0.105000)(0.079,0.105000)(0.080,0.105000)(0.081,0.104000)(0.082,0.104000)(0.083,0.104000)(0.084,0.104000)(0.085,0.104000)(0.086,0.104000)(0.087,0.104000)(0.088,0.104000)(0.089,0.104000)(0.090,0.104000)(0.091,0.104000)(0.092,0.104000)(0.093,0.104000)(0.094,0.104000)(0.095,0.104000)(0.096,0.104000)(0.097,0.104000)(0.098,0.104000)(0.099,0.103000)(0.100,0.103000)(0.200,0.101000)(0.300,0.100000)(0.400,0.100000)(0.500,0.100000)(0.600,0.100000)(0.700,0.100000)(0.800,0.100000)(0.900,0.100000)(1.000,0.099000)
};
	    \addlegendentry{$K$};
    \path[name path=top] (axis cs:0,1) -- (axis cs:.1,1);
    \addplot[thick,color=red,fill=red,fill opacity=0.3] fill between [of=g and top];
    \addplot[thick,color=blue,fill=blue,fill opacity=0.3] fill between [of=f and top];

    \node[red!40!black] at (axis cs:.042,0.85) {Classification possible};
    \node[red!40!black] at (axis cs:.042,0.75) {with $K_{\rm sub}$ and $K$};
    \node[blue] at (axis cs:.055,0.22) {Classification possible \emph{only} with $K$};
    \node[black] at (axis cs:.06,0.04) {Classification impossible};
\end{axis}
\end{tikzpicture}
\caption{Phase transition diagram of spectral clustering for puncturing matrix $K$ with $\varepsilon_S=1$ and $\varepsilon_B\equiv\varepsilon$ versus subsampling $K_{\rm sub}\in\mathbb C^{n\varepsilon\times n\varepsilon}$. Here for $x_i\sim \frac12\mathcal{CN}(\mu,I_p)+\frac12\mathcal{CN}(-\mu,I_p)$, and $n/p=100$ in the large $n,p$ limit. Solid and dashed lines indicate theoretical phase transitions. \textbf{\textit{The puncturing approach largely overtakes the subsampling method.}} }
\label{fig:subsampling}
\end{figure}

To cope with these limitations, the present article introduces a new random data sparsification method which trades off storage and computational cost reduction against performance. The proposed \emph{two-way puncturing} approach consists in random Bernoulli deletions of entries (i) of the data matrix $X=[x_1,\ldots,x_n]\in\mathbb C^{p\times n}$ (the indices of non-zero entries differing across data) and (ii) of the Gram (sample covariance $\frac1nXX^\H$ or kernel $\frac1pX^\H X$) matrix, generically resulting in the kernel matrix model
\begin{align}
    \label{eq:K}
    K &= \left\{\frac1p(X\odot S)^\H(X\odot S)\right\} \odot B \in \mathbb{C}^{n\times n}
\end{align}
for random independent Bernoulli $S\in\{0,1\}^{p\times n}$ and (symmetric) $B\in\{0,1\}^{n\times n}$, with respective parameters $\varepsilon_S$ and $\varepsilon_B\in(0,1]$. Small values of $\varepsilon_S$ reduce the storage size of $X$ and the cost of the inner-product evaluation $x_i^\H x_j$, while small values of $\varepsilon_B$ reduce the number of inner-product calculus in $K$ and the subsequent processing of the sparsified matrix $K$. The approach follows after our preliminary work \citep{zarrouk2020performance}, restricted to $S=1_p1_n^{\sf T}$ (or equivalently $\varepsilon_S=1$) and to a simpler model for $X$, which already revealed that, contrary to intuition, the puncturing procedure in general \emph{does not affect the structure} of the estimated eigenvectors (thus principal components in PCA or data classes in clustering). This conclusion still holds true here. More surprisingly, the analysis also demonstrates that there exist well-defined regimes -- in terms of the ratio $p/n$ and puncturing intensities $\varepsilon_S$ and $\varepsilon_B$ -- for which the \emph{PCA performance is virtually unaltered}. In particular, for equivalent levels of sparsity (in terms of resulting computational costs), we confirm here the finding of \citep{zarrouk2020performance} according to which the performance of PCA or spectral clustering on $K$ largely overtakes the performance of the possibly more natural subsampling alternative.\footnote{Subsampling consists here in performing PCA or spectral clustering on $n/\varepsilon$ subsets of the data, each of size $\varepsilon n$, for some $\varepsilon \in (0,1]$ a multiple of $1/n$, before merging the $n/\varepsilon$ results (which for simplicity we assume here comes at no cost).} This result is recalled in Figure~\ref{fig:subsampling} for $\varepsilon_S=1$ and $\varepsilon_B\equiv \varepsilon$.

\medskip

Our main findings may be summarized as follows:
\begin{enumerate}
    \item for data $x_i$ arising from a Gaussian mixture model $\sum_{\ell=1}^k\pi_\ell \mathcal N(\mu_\ell,I_n)$ (resp., a Gaussian measure $\mathcal N(0,C)$ with $C=I_p+R$ and $R$ of low rank), we show that $K$ has a limiting eigenvalue distribution following a \emph{variation of the popular Mar\u{c}enko-Pastur and semi-circle laws}; upon conditions on the eigenvalues of the matrix $\{\sqrt{\pi_i\pi_j}\mu_i^\T\mu_j\}_{i,j=1}^k$ (resp., of the matrix $R$), a phase transition phenomenon occurs beyond which some eigenvalues of $K$ isolate, and their associated eigenvectors correlate to the population eigenvectors;
    \item the quantities $p/n$, $\varepsilon_S$, and $\varepsilon_B$ modulate the storage-and-computational cost versus (PCA or spectral clustering) performance trade-off; in particular, for small $\varepsilon_S,\varepsilon_B$, the performance only depends on $\varepsilon_S^2\varepsilon_B\frac{p}{n}$;
    \item for small $p/n$ ratios (i.e., for huge amounts of data), the performance of PCA and spectral clustering \emph{plateaus} for a large range of values of $\varepsilon_B$ (with $\varepsilon_S$ fixed), before suffering a sharp avalanche phenomenon for $\varepsilon_B$ below a certain threshold: this in particular indicates that \emph{intensive puncturing (and thus complexity and storage reduction) almost comes for free} in this regime;
    \item simulations on Fashion-MNIST and BigGAN generated images qualitatively (and partially quantitatively) confirm our theoretical findings, justifying the possibility to drastically reduce computational cost with virtually no impairment on classification performance.
\end{enumerate}

\noindent{\bf Supplementary material and codes.} The proofs of our main results are deferred to the supplementary material. All codes to reproduce our figures are available in the gitlab repository {\scriptsize \url{https://gricad-gitlab.univ-grenoble-alpes.fr/chatelaf/two-way-kernel-matrix-puncturing}}. 

\section{The two-way puncturing model}
\label{sec:model}

Before relating our study to principal component analysis and spectral clustering, we first formalize the model under study in a generic (and thus abstract) manner.

\subsection{Abstract model}

Let $X\in\mathbb C^{p\times n}$ be a random matrix satisfying the following assumptions.\footnote{All results are provided in $\mathbb C$ but are equally valid in $\mathbb R$.}
\begin{assumption}[Data model]
\label{ass:model}
\begin{align*}
    X &= Z + P
\end{align*}
in which $Z_{ij}\sim \mathcal{CN}(0,1)$ are independent, and $P\in\mathbb C^{p\times n}$ is a rank-$k$ matrix for some integer $k$.
\end{assumption}

Also define the binary \emph{puncturing} matrices $S\in\{0,1\}^{p\times n}$ and $B\in\{0,1\}^{n\times n}$ as follows.
\begin{assumption}[Puncturing matrices]
\label{ass:puncturing}
Let
\begin{itemize}[noitemsep,nolistsep]
    \item $S_{ij}\in\{0,1\}$ be Bernoulli random variables with mean $\varepsilon_S$, independent across $i,j$;
    \item $B_{ij}=B_{ji}\in\{0,1\}$ be Bernoulli random variables with mean $\varepsilon_B$, independent across $i>j$;
    \item $B_{ii}=b\in\{0,1\}$ be deterministic and fixed.
\end{itemize}
Besides, matrices $S$, $B$, and $X$ are mutually independent.
\end{assumption}

Our objective is to study the spectral properties of the random matrix model \eqref{eq:K}.
Specifically, we determine the limiting spectrum as well as the existence and characterization of isolated eigenvalues (i.e., away from the limiting spectrum and referred to as \emph{spikes}) and their associated eigenvectors, in the limit of large $p,n$. To this end, the following growth rate assumptions are requested.

\begin{assumption}[Large $p,n$ asymptotics]
\label{ass:asymptotics}
As $n\to\infty$,
\begin{align*}
    p/n &\to c_0\in(0,\infty)
\end{align*}
and there exists a decomposition $P=LV^\H$ of $P$ with $V\in\mathbb C^{n\times k}$ isometric (i.e., $V^\H V=I_k$) and
\begin{align*}
    \frac1nL^\H L &\to \mathcal L
\end{align*}
for some deterministic matrix $\mathcal L \in\mathbb C^{k\times k}$. In particular, the eigenvalues of $\mathcal L$ are the limiting $k$ non-trivial eigenvalues of 
$\frac1nP^\H P$. Besides,
\begin{align*}
    \limsup_n\max_{ \substack{1\leq i\leq n \\ 1\leq j\leq k}}\{\sqrt nV_{ij}^2\}=0.
\end{align*}
\end{assumption}
The condition $p/n\to c_0\in (0,\infty)$ translates the practical fact that both the dimension and number of data are large and commensurable. The convergence $(1/n)L^\H L\to \mathcal L$ with $P=LV^\H$ is merely technical: the decomposition $P=LV^\H$ can always be ensured by singular value decomposition, and the convergence to $\mathcal L$ is mostly for technical convenience. In effect, the only stringent condition is that $\limsup_n\max_{i,j} \sqrt n V_{ij}^2=0$: while naturally satisfied for spectral clustering (the $V_{ij}$'s are the normalized binary class indicators), for PCA this demands that the principal components be \emph{delocalized}, i.e., not sparse.

\subsection{PCA and spectral clustering}
\label{sec:PCA_spect_clust}
The model \eqref{eq:K} specializes to principal component analysis and spectral clustering.

\noindent {\bf Spectral clustering}. Letting $P=MJ^\T$, where $M=[\mu_1,\ldots,\mu_k]\in\mathbb C^{p\times k}$ and $J=[j_1,\ldots,j_k]\in\mathbb \{0,1\}^{n\times k}$ with $[j_\ell]_i=\delta_{\{\mathbb E[x_i]=\mu_\ell\}}$ for some $n_1,\ldots,n_k$, $X$ models a $k$-class Gaussian mixture model with $x_i\sim \sum_{a=1}^k\pi_a\mathcal{CN}(\mu_a,I_p)$ and $n_a/n\to \pi_a$ almost surely as $n\to\infty$. Further assuming that
\begin{align*}
    n_\ell/n &\to \pi_\ell = [\pi]_\ell \in(0,\infty) \\
    \mathcal D_\pi^{\frac12}M^\H M\mathcal D_\pi^{\frac12} &\to \mathcal M
\end{align*}
where $\mathcal D_\pi={\rm diag}(\{\pi_i\}_{i=1}^k)$, we get that $P=(MD_n^{\frac12})(JD_n^{-\frac12})^\H$ with $D_n={\rm diag}(\{n_i\}_{i=1}^k)$, for which
\begin{align*}
    (JD_n^{-\frac12})^\T(JD_n^{-\frac12}) = I_k,\quad
    \frac1n (MD_n^{\frac12})^\H(MD_n^{\frac12}) \to \mathcal M
\end{align*}
thereby satisfying Assumptions~\ref{ass:model}--\ref{ass:asymptotics}, for 
$\mathcal L=\mathcal M$. Under this setting, $\frac1pX^\H X$ is (the elementary version of) a kernel random matrix used in machine learning as the base ingredient for kernel-based classification methods. In particular, the eigenvectors associated with the dominant eigenvalues of $\frac1pX^\H X$ are the base elements of the popular \emph{(kernel) spectral clustering} algorithm \cite{von2007tutorial}.

\medskip

\noindent {\bf Principal component analysis}. Letting instead $P=\tilde Z A^\H$ with $A\in\mathbb C^{n\times k}$ deterministic and $\tilde Z\in\mathbb C^{p\times k}$ random with i.i.d.~$\mathcal {CN}(0,1)$ entries, independent of $Z$, we get
\begin{align*}
    X^\H &= \begin{bmatrix} I_n & A \end{bmatrix} \begin{bmatrix} Z^\H \\ \tilde Z^\H\end{bmatrix}
\end{align*}
which is a matrix with $\mathcal {CN}(0,I_n+AA^\H)$ independent columns, so that $\frac1pX^\H X$ is a sample covariance matrix for the $p$ \emph{rows}\footnote{One must be careful here that standard notations of $n$ and $p$ are reversed under this setting.} of $X$ of dimension $n$; the dominant eigenvectors of $\frac1pX^\H X$ are therefore the principal components of the popular \emph{principal component analysis} method. Further requesting $A$ to have spectral decomposition $A=USV^\H$, where $S\in\mathbb R_+^{k\times k}$ satisfies $S^\H S\to \mathcal S$ deterministic, one gets that $P = (\tilde Z US)V^\H$ with $V^\H V=I_k$ and
\begin{align*}
    \frac1n (\tilde Z US)^\H(\tilde Z US) \to \mathcal S
\end{align*}
again satisfying Assumption~\ref{ass:asymptotics} for $\mathcal L=\mathcal S$.

\section{Main results}
\label{sec:results}

As per standard random matrix methods, the technical approach to study the limiting spectrum of $K$ consists in characterizing the \emph{resolvent} matrix
\begin{align*}
    Q(z) &= (K-zI_n)^{-1}
\end{align*}
defined for $z\in\mathbb C\setminus\{\lambda_i\}_{i=1}^n$ with $\lambda_i$ the eigenvalues of $K$. Specifically, the \emph{spectral measure} $\nu_n \equiv \frac1n\sum_{i=1}^n \delta_{\lambda_i}$ of $K$ relates to the \emph{Stieltjes transform} $m_n(z)\equiv\int (t-z)^{-1}\nu_n(dt)=\frac1n\tr Q(z)$, while the \emph{eigenvector} $\hat u_i\in\mathbb C^n$ associated to eigenvalue $\lambda_i(K)$ relates to the Cauchy-integral $\hat u_i\hat u_i^\H=\frac{-1}{2\pi\imath}\oint_{\Gamma_{\lambda_i}}Q(z)dz$ for $\Gamma_{\lambda_i}$ a small positively oriented complex contour circling around $\lambda_i$ only.

\subsection{Limiting spectral behavior}

Our core technical result provides a said \emph{deterministic equivalent} for the random matrix $Q(z)$, from which the limiting behavior of the eigenvalues and eigenvectors of $K$ follows.

\begin{theorem}[Deterministic equivalent for $Q$]
    \label{th:barQ}
    Under Assumptions~\ref{ass:model}--\ref{ass:asymptotics}, let $z\in\mathbb C$ be away from the limsup of the union of the supports of $\nu_1,\nu_2,\ldots$. Then, as $n\to\infty$,
    \begin{align*}
        Q(z) &\leftrightarrow m(z) \left[I_n + \frac{c_0^{-1}\varepsilon_S^2\varepsilon_B m(z)}{1+\varepsilon_B\varepsilon_Sc_0^{-1}m(z)}V\mathcal LV^\H\right]^{-1}
    \end{align*}
    where $m(\cdot)$ is the unique Stieltjes transform solution to
    \begin{align*}
    z = \varepsilon_Sb - \frac1{m(z)} - c_0^{-1}\varepsilon_B\varepsilon_S^2m(z)+\frac{c_0^{-2}\varepsilon_B^3\varepsilon_S^3m(z)^2}{1+c_0^{-1}\varepsilon_B\varepsilon_S m(z)}
\end{align*}
and the notation $A\leftrightarrow B$ indicates that, for any linear functional $u:\mathbb C^{n\times n}\to \mathbb R$ of bounded infinity norm, $u(A-B)\to 0$ almost surely as $n\to\infty$.
\end{theorem}

One must understand the theorem as follows: since $Q(z)$ encapsulates the structural spectral information about $K$, this information is fully determined (in the large $n,p$ limit)
\begin{itemize}
    \item[(i)] by the scalars $\varepsilon_S$, $\varepsilon_B$, $c_0$ and $b$; these mostly impact the \emph{shape} of the limiting spectrum in defining $m(\cdot)$) and modulate the ``noise level'' of the eigenvectors (from the factor preceding $V\mathcal LV^\H$ in the expression of $Q(\cdot)$);
    \item[(ii)] by the rank-$k$ matrix $V\mathcal LV^\H$; this matrix defines the ``average'' behavior of the dominant eigenvectors of $K$: these eigenvectors are simply ``isotropic noisy versions'' of linear combinations of the columns of $V$. That is, mapped to the applications in Section~\ref{sec:PCA_spect_clust}, noisy versions of either the class canonical vectors $j_a$'s or of the genuine PCA vector.
\end{itemize}
As an immediate -- and possibly quite surprising -- consequence, the dominant eigenvectors of $K$ are, up to extra \emph{homogeneous noise}, the same as those of $P^\H P=\mathbb E[\frac1pX^\H X]-I_n$. The proposed two-way puncturing algorithm therefore \emph{does not affect spectral algorithms} as the structure of the retrieved eigenvectors is maintained.

\medskip

\begin{figure}[t]
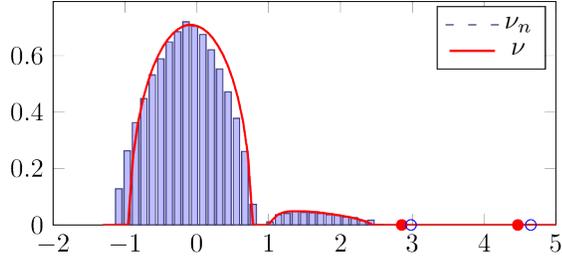

    \centering
    \includestandalone{fig_dist_hist}
    \vspace{-3mm}\\
    \caption{Eigenvalue distribution $\nu_n$ of $K$ versus limit measure $\nu$, for $p=200$, $n=4\,000$, $x_i\sim .4 \mathcal N(\mu_1,I_p)+.6\mathcal N(\mu_2,I_p)$ for $[\mu_1^\T,\mu_2^\T]^\T \sim\mathcal N(0, \frac1p\left[\begin{smallmatrix} 10 & 5.5 \\ 5.5 & 15\end{smallmatrix}\right]\otimes I_p)$; $\varepsilon_S=.2$, $\varepsilon_B=.4$, $b=1$. Sample vs theoretical spikes in blue vs red circles. \textbf{\textit{The two ``humps'' remind the semi-circular and Mar\u{c}enko-Pastur laws.}} }
    \label{fig:spectrum}
\end{figure}

Let us now quantify these so far qualitative statements. As a first corollary of Theorem~\ref{th:barQ}, with probability one,
\begin{align*}
    \frac1n\tr Q(z) &\equiv m_n(z) \to m(z)
\end{align*}
which implies, according to random matrix theory, that
\begin{align*}
    \nu_n \equiv \frac1n\sum_{i=1}^n\delta_{\lambda_i} \to \nu
\end{align*}
almost surely, where $\nu$ is the unique probability measure having Stieltjes transform $m(z)$ (i.e., $m(z)=\int (t-z)^{-1}\nu(dt)$). It thus suffices to solve the defining equation for $m(z)$ in Theorem~\ref{th:barQ} to estimate the limiting spectral distribution $\nu$ of $K$.\footnote{The measure $\nu$ is practically retrieved from $m(\cdot)$ by using the inverse formula $\nu(dt)=\lim_{y\downarrow 0}\frac1\pi \Im[m(t+\imath y)]dt$.} Figure~\ref{fig:spectrum} indeed confirms the correspondence between the empirical (finite $n,p$) spectrum $\nu_n$ of $K$ versus the estimated limit $\nu$.


\begin{remark}[Sitting between Mar\u{c}enko-Pastur and Wigner]
    Not surprisingly, when $\varepsilon_B=1$ and $b=1$, $K=\frac1p(X\odot S)^\H(X\odot S)$ with $X\odot S$ a matrix with i.i.d.~entries of zero mean and variance $\varepsilon_S^2$, so that $\nu$ falls back onto the popular Mar\u{c}enko-Pastur distribution \cite{marvcenko1967distribution} (up to an $\varepsilon_S$ scale). Precisely,
    for $z'=z/\varepsilon_S$ and $\tilde m(z)=\int (t/\varepsilon_S-z)^{-1}\nu(dt)$ (i.e., the Stieltjes transform of the limiting measure of the $\lambda_i/\varepsilon_S$), the canonical equation of $m(z)$ in Theorem~\ref{th:barQ} becomes
    \begin{align*}
        z' &= 1 - \frac1{\tilde m(z')} - \frac{c_0^{-1}\tilde m(z')}{1+c_0^{-1}\tilde m(z')}
    \end{align*}
    which is the defining Stieltjes transform equation of the Mar\u{c}enko-Pastur law.
    The more interesting small $\varepsilon_B$ setting is treated in Section~\ref{sec:small_eS_eB_limit} and gives rise to a Wigner semi-circle limit instead \cite{wigner1958distribution}. As such, through the values $\varepsilon_S,\varepsilon_B$, the limiting spectral measure $\nu$ continuously moves from the Mar\u{c}enko-Pastur to the Wigner semi-circle laws. Figure~\ref{fig:spectrum} illustrates this observation: the shape of $\nu$ is simultaneously reminiscent of both laws.
\end{remark}

\subsection{Phase transition and dominant eigenvectors}
\label{sec:phase_transition}
The limiting Stieltjes transform $m(z)$ determines the ``macroscopic'' behavior of the spectrum $\nu_n$ of $K$, but does not provide the position of its isolated eigenvalues and even less the shape of the associated eigenvectors. To this end, a deeper investigation of the deterministic equivalent of $Q(z)$ is needed. Our next result provides this analysis.

\begin{theorem}[Phase transition, isolated eigenvalues and eigenvectors]
    \label{th:spikes}
    Define the functions
    \begin{align*}
        F(t)&=t^4+\frac{2}{\varepsilon_S}t^3+\frac{1}{\varepsilon_S^2}\left( 1 - \frac{c_0}{\varepsilon_B}\right)t^2 - \frac{ 2c_0}{\varepsilon_S^3}t-\frac{c_0}{\varepsilon_S^4} \\
        G(t)&= \varepsilon_Sb+c_0^{-1}\varepsilon_B\varepsilon_S(1+\varepsilon_St)+\frac{\varepsilon_S}{1+\varepsilon_St}+\frac{\varepsilon_B}{t(1+\varepsilon_St)}
    \end{align*}
    and $\Gamma\in\mathbb R$ be the largest real solution to $F(\Gamma)=0$.
    Further denote $\ell_1>\ldots>\ell_{\bar k}$ the $\bar k\leq k$ \emph{distinct} eigenvalues of $\mathcal L$ of respective multiplicities $L_1,\ldots,L_{\bar k}$, and $\Pi_1,\ldots,\Pi_{\bar k}\in\mathbb R^{k\times k}$ the projectors on their respective associated eigenspaces. Similarly denote $(\lambda_1,\hat v_1),\ldots,(\lambda_n,\hat v_n)$ the eigenvalue-eigenvector pairs of $K$ in descending order and gather the first $k$ eigenvectors under the isometric matrices $\hat{\mathcal V}_1=[\hat v_1,\ldots,\hat v_{L_1}]$ up to $\hat{\mathcal V}_{\bar k}=[\hat v_{k-L_{\bar k}+1},\ldots,\hat v_k]$.

    Then, for $i\in\{1,\ldots,\bar k\}$ and for all $j\in\{L_1+\ldots+L_{i-1}+1,\ldots,L_1+\ldots+L_i\}$,
    \begin{align*}
        \lambda_j &\to \rho_i \equiv \left\{
        \begin{array}{ll}
             G(\ell_i) &,~\textmd{if}~\ell_i>\Gamma  \\
             G(\Gamma) &,~\textmd{if}~\ell_i\leq \Gamma
        \end{array}\right.
        \end{align*}
        almost surely, and
        \begin{align*}
        \hat{\mathcal V}_i\hat{\mathcal V}_i^\H \leftrightarrow \zeta_i V\Pi_i V^\H,~\textmd{for}~\zeta_i = \left\{
        \begin{array}{ll}
        \frac{F(\ell_i)\varepsilon_S^3}{\ell_i(1+\varepsilon_S\ell_i)^3} &,~\ell_i>\Gamma \\
        0&,~\ell_i\leq\Gamma
        \end{array}
        \right.
    \end{align*}
    with the notation `$\leftrightarrow$' introduced in Theorem~\ref{th:barQ}. In particular, if the $\ell_i$'s have unit multiplicities with associated population eigenvectors $v_i$, then
    \begin{align*}
        |v_i^\H \hat v_i|^2 \to \zeta_i,\quad i=1,\ldots,k.
    \end{align*}
\end{theorem}

\begin{figure}[t]
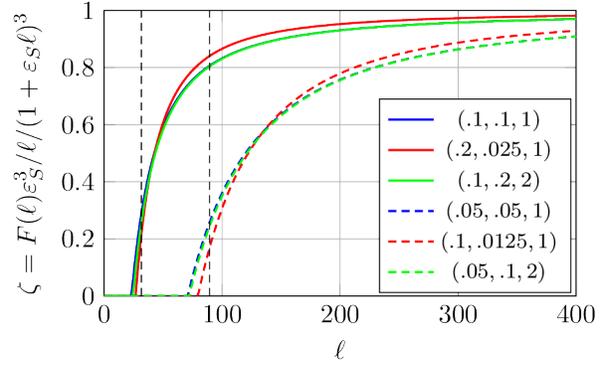

    \centering
    \includestandalone{fig_eigvect_alignment}
    \vspace{-3mm}\\
    \caption{Illustration of Theorem~\ref{th:spikes}: asymptotic sample-population eigenvector alignment for $\mathcal L=\ell \in\mathbb R$, as a function of the ``information strength'' $\ell$. Various values of $(\varepsilon_S,\varepsilon_B,c_0)$ indicated in legend. Black dashed lines indicate the limiting (small $\varepsilon_S,\varepsilon_B$) phase transition threshold $\Gamma=(\varepsilon_S^2\varepsilon_Bc_0^{-1})^{-\frac12}$. \textbf{\textit{As $\varepsilon_S,\varepsilon_B\to 0$, performance curves coincide when $\varepsilon_B\varepsilon_S^2c_0^{-1}$ is constant (plain versus dashed set of curves).}}}
    \label{fig:alignment}
\end{figure}

To best understand the theorem, suppose that $P=lv^\H$ is a rank-$1$ matrix with $\|v\|^2=1$ and $\|l\|^2/n=\ell$. Then, if $\ell>\Gamma$, with $\Gamma$ the largest solution to $F(\Gamma)=0$ -- this threshold \emph{only depending on $\varepsilon_S$, $\varepsilon_B$ and $c_0$} --, the spectrum of $K$ exhibits an isolated eigenvalue $\lambda$, the eigenvector $\hat v$ of which \emph{aligns} to $v$: i.e., $|\hat v^\H v|^2\to \zeta>0$. Otherwise, if $\ell<\Gamma$, the largest eigenvalue $\lambda$ of $K$ remains ``stuck'' in the limiting bulk of eigenvalues of $K$ and $|\hat v^\H v|^2\to 0$ (i.e., the eigenvector $\hat v$ does not carry any information on $v$: PCA and spectral clustering both fail in this scenario). Figure~\ref{fig:alignment} illustrates the limiting (squared) alignment $\zeta$ as a function of $\ell$.

In the more general setting where $P$ is a rank-$k$ matrix, possibly with multiplicities, the theorem specifies the conditions on $\varepsilon_B$, $\varepsilon_S$ and $c_0$ under which the dominant eigenvectors of $K$ remain correlated (and to which extent) to the population eigenspaces. This characterization is of tremendous importance to assess the exact performance of PCA and spectral clustering under the double-puncturing cost reduction. Figure~\ref{fig:alignment} illustrates Theorem~\ref{th:spikes} in a clustering setting.


\medskip

An important quantity of Theorem~\ref{th:spikes} is the function $F$, which intervenes both to establish the condition under which informative isolated eigenvalues are found in the spectrum of $K$, thereby defining the \emph{phase transition} threshold for the population eigenvalue $\ell_i$ (through $F(\ell_i)=0$), and to evaluate the corresponding empirical eigenvector(s) quality through $\zeta_i=F(\ell_i)\varepsilon_S^3/(\ell_i(1+\varepsilon_S\ell_i)^3)$ (which is zero right at the phase transition threshold). The phase transition determines which values of the tuple $(\varepsilon_S,\varepsilon_B,c_0,\ell_i)$ coincide with the emergence of an isolated eigenvalue in the spectrum of $K$ associated to the population eigenvalue $\ell_i$, and thus to the actual feasibility of PCA or spectral clustering.

\begin{figure}[t]
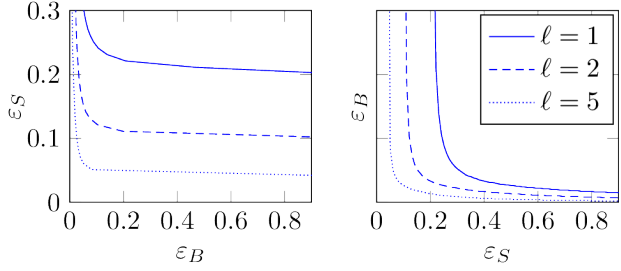

    \centering
    \hspace*{-.5cm}\begin{tabular}{cc}
    \includestandalone{fig_phase_trans_S}
    &
    \includestandalone{fig_phase_trans_B}
    \end{tabular}
    \vspace{-3mm}\\
    \caption{Phase transition curves $F(\ell)=0$ for $\mathcal L=\ell\in\mathbb R$ and varying values of $\ell$, for $c_0=.05$. Above each phase transition curve, a spike eigenvalue is found away from the support of $\nu$. \textbf{\textit{For large $\ell$, a wide range of $\varepsilon_B$'s (resp.\@ $\varepsilon_S$) is admissible at virtually no performance loss. Here, also, sparser $B$ matrices are more effective than sparser $S$ matrices.}} }
    \label{fig:equi-perf}
\end{figure}

Assume now that $c_0\ll 1$ (i.e., $n\gg p$) and that $\varepsilon_B$ and $\ell_i$ are kept fixed and away from zero. Then, in the expression of $F(\ell_i)$, $1\gg c_0/\varepsilon_B$ so that, in the first order, $F(\ell_i)$ is independent of $\varepsilon_B$. This quite importantly implies that the ``function'' $\varepsilon_S:\varepsilon_B\mapsto \varepsilon_S(\varepsilon_B)$ such that $F(\ell_i)=0$ is mostly flat for a range of non-small values of $\varepsilon_B$. This behavior is confirmed in Figure~\ref{fig:equi-perf} (left display). Also, since $\zeta_i$ would also marginally depend on $\varepsilon_B$, the eigenvector quality is also the same for a wide range of $\varepsilon_B$. The major consequence of this remark is that, for $c_0\ll 1$, $\varepsilon_B$ can be taken quite small without affecting the quality of the dominant eigenvectors: \emph{puncturing through $B$ does not affect the PCA or spectral clustering performance and thus almost comes for free!}

\enlargethispage*{5mm}

Conversely, still for $c_0\ll 1$, for $\varepsilon_S$ fixed and away from zero, we find that, at the phase transition,
\begin{align*}
    \varepsilon_B &\simeq {c_0}/{(1+\varepsilon_S\ell_i)^2}.
\end{align*}
As such, the reverse function $\varepsilon_B(\varepsilon_S)$ is quite different from $\varepsilon_S(\varepsilon_B)$: it mostly behaves as $1/\varepsilon_S^2$ so that, in order not to loose performance, increased sparsification through $S$ must come along with reduced sparsification through $B$.

\medskip

Of utmost interest though is the case where $c_0$ and $\ell_i$ are fixed (although, as we will see, $\ell_i^2/c_0$ must be large), and where both Bernoulli parameters $\varepsilon_B$ and $\varepsilon_S$ assume small values. This scenario is all the more relevant that Theorems~\ref{th:barQ}--\ref{th:spikes} and their corollaries take on simple and intuitive forms. This setting is discussed next.


\subsection{Small $\varepsilon_B$, $\varepsilon_S$ limit}
\label{sec:small_eS_eB_limit}
Letting $z'=\sqrt{{c_0}/{\varepsilon_B\varepsilon_S^2}}(z-\varepsilon_Sb)$, we obtain, in the limit of small $\varepsilon_B$ and $\varepsilon_S$, that
\begin{align*}
    z' &= -1/{m_0(z')}-m_0(z')+o\left( ({\varepsilon_B\varepsilon_S^2c_0^{-1}})^{\frac12} m_0(z')\right)
\end{align*}
with $m_0(z)=({c_0^{-1}\varepsilon_B\varepsilon_S^2})^{\frac12}m(({c_0^{-1}\varepsilon_B\varepsilon_S^2})^{\frac12}z+\varepsilon_Sb)$, i.e., for $\nu$ the measure associated to $m(z)$, $m_0$ is the Stieltjes transform of the measure $\nu(t/({c_0^{-1}\varepsilon_B\varepsilon_S^2})^{\frac12})$.

This is the defining equation of Wigner's semi-circle law \citep{wigner1958distribution} centered at $\varepsilon_Sb$ and with edges $\varepsilon_Sb\pm 2({c_0^{-1}\varepsilon_B\varepsilon_S^2})^{\frac12}$.

Similarly, assuming $\ell_i>\Gamma$, and letting $\rho_i'=(\rho_i-\varepsilon_Sb)/{c_0^{-1}\varepsilon_B\varepsilon_S^2}^{\frac12}$ and $\ell_i'=\ell_i({\varepsilon_B\varepsilon_S^2c_0^{-1}})^{\frac12}$, we find, after first order Taylor expansion, the spike equation $m_0(\rho'_i) = -1/{\ell_i'} + o( ({\varepsilon_B\varepsilon_S^2c_0^{-1}})^{\frac12})$, or equivalently
\begin{align*}
    \rho_i' &= \ell_i'+1/{\ell_i'} + o\left( ({\varepsilon_B\varepsilon_S^2c_0^{-1}})^{\frac12}\right)
\end{align*}
which is the classically known isolated eigenvalue from the deformed Wigner random matrix \citep[Chapter~2.2]{pastur2011eigenvalue}. The scaling of $\ell_i$ into $\ell_i'$ importantly indicates that, for a non-trivial spike to emerge, the eigenvalue $\ell_i$ of $\mathcal L$ must scale like $O(({c_0\varepsilon_B^{-1}\varepsilon_S^{-2}})^{\frac12})$.

In practical terms, these results show that (i) for spectral clustering to be feasible (but non-trivial), the inter-class distance $\|\mu_a-\mu_b\|^2$ must scale like $\sqrt{c_0/(\varepsilon_B\varepsilon_S^2)}$, and (ii) for PCA, the eigenvalues of the principal components must scale like $\sqrt{c_0/(\varepsilon_B\varepsilon_S^2)}$.

\medskip

As for the alignment of eigenspaces, it is given by
\begin{align*}
    \hat{\mathcal U}_i\hat{\mathcal U}_i' \leftrightarrow \left( 1-1/{(\ell_i')^2} + o\left(({\varepsilon_B\varepsilon_S^2c_0^{-1}})^{\frac12}\right) \right) V \Pi_i V^\H
\end{align*}
which, again, is a classical result in the deformed Wigner random matrix model. Setting the alignment to zero, this result also provides a much simpler value for the phase transition threshold $\ell_i=\Gamma$ of Theorem~\ref{th:spikes} (in the limit of small $\varepsilon_S,\varepsilon_B$) which corresponds to $\ell_i'\simeq 1$, or equivalently
\begin{align*}
    \Gamma \simeq 1/{({\varepsilon_B\varepsilon_S^2c_0^{-1}})^{\frac12}}.
\end{align*}

\medskip

\begin{remark}[Trading off $\varepsilon_B$, $\varepsilon_S$ and $c_0$]
\label{rem:eB_eS2_c0}
As a consequence of the results above, it appears that, for small values of $\varepsilon_B$, $\varepsilon_S$ and $c_0^{-1}$, the spectral behavior (eigenvalues and eigenvectors) of $K$ is unaltered so long that $\varepsilon_B\varepsilon_S^2c_0^{-1}$ is constant. For instance, doubling $n$ is equivalent to doubling $\varepsilon_B$ or multiplying $\varepsilon_S$ by $\sqrt 2$. This is confirmed by Figure~\ref{fig:alignment} in which the two sets of plain or dashed curves, corresponding to constant $\varepsilon_B\varepsilon_S^2c_0^{-1}$, almost coincide.

It is important to further note that, unlike $\varepsilon_B$, $\varepsilon_S$ is squared in the expression $\varepsilon_B\varepsilon_S^2c_0^{-1}$ due to the fact that, denoting $S=[s_1,\ldots,s_n]$, the inner products $(x_i\odot s_i)^\H(x_j\odot s_j)$, for all $i\neq j$, involve on average $\varepsilon_S^2$ terms (since $\frac1p\mathbb E[s_i^\T s_j]=\varepsilon_S^2$).
\end{remark}

One must be careful not to confuse the findings of Section~\ref{sec:phase_transition} on \emph{non-small $\varepsilon_B$} according to which $\varepsilon_B\in(0,1]$ has a marginal impact on performance (and thus that intensive puncturing comes for free), to the present results which on the opposite indicate that \emph{for small $\varepsilon_B$}, more intensive puncturing decreases the performance. Both regimes are very different as Figure~\ref{fig:equi-perf} clearly indicates.

\section{Practical consequences: the storage/complexity performance trade-off}

The main interest of the two-way puncturing approach lies in its effective computational and storage cost reductions, while maintaining high performance levels. As a follow-up of Remark~\ref{rem:eB_eS2_c0}, puncturing through the matrix $S$ can be traded off by puncturing through $B$, and vice-versa, with, we will see, varying effects on storage and computational costs.

\subsection{Storage and computation costs}

\paragraph{Computing $K$.} For $B_{ij}=1$, evaluating $K_{ij}$ comes at average cost of $\mathbb E[\sum_{\ell=1}^p S_{i\ell}S_{\ell j}]=\varepsilon_S^2$ products. As a result, the whole matrix $K$, with an average $\sum_{i,j=1}^n\mathbb E[B_{ij}]=\varepsilon_Bn^2$ (if $b=1$, and $\varepsilon_B(n-1)^2$ if $b=0$) non-zero entries, has $O(n^2p\varepsilon_S^2\varepsilon_B)$ theoretical computation cost.

\paragraph{Storage data.} In terms of storage, if one wishes to maintain the data information $X\odot S$ for further (non-kernel related) use, the net gain is a factor $\varepsilon_S$ on average (for a net storage of $\varepsilon_Spn$ values). If instead only the matrix $K$ is of relevance for future use, then the storage is restricted to $\varepsilon_Bn(n-1)/2+n$ values when $b=1$ (accounting for symmetry) or $\varepsilon_Bn(n-1)/2$ values when $b=0$.

\paragraph{Spectral methods.} When it comes to spectral methods (PCA or spectral clustering), one needs to retrieve the (few) dominant eigenvectors of $K$. Using a power method on $K$ to sequentially iterate over each eigenvector is in general optimal and comes at a cost of $O(\varepsilon_Bn^2)$, where the $O(\cdot)$ notation encompasses the number of iterations required for convergence (which depends on the spectral gap between isolated eigenvalues and thus does not scale with $n$ in our setting). This is a gain of order $\varepsilon_B$ over no puncturing.

Yet, when $p\ll n$, to evaluate the dominant eigenpairs of $X^\H X$, it is more efficient in practice to proceed to a singular value decomposition of the $n\times p$ matrix $X^\H$, again via a power method. When operating the Hadamard product with $B$ though, this strategy cannot be put in place as $X^\H X \odot B$ is in general of full rank $n$. It is thus in this case beneficial to divert the sparsity into letting $\varepsilon_B=1$ and $\varepsilon_S\ll 1$ so to be able to run a singular vector decomposition over the very sparse matrix $(X\odot S)^\H$.


\begin{remark}[Cache issues]
    \label{rem:cache}
    The computational costs reported in this section are provided in terms of net number of product operations, irrespective of computer architecture or implementation. But computing the entries of the Gram matrix $X^\H X$ can be advantageously performed ``block-wise'' by caching vectors in sequences of blocks and computing the corresponding subblocks of $X^\H X$. This powerful trick cannot be performed on the two-way punctured matrix $K$ which, due to the randomness in $S$ and $B$, is not organized in blocks.
    In practice, we observed that the cost of systematically retrieving the $x_i$'s by pairs from remote memory is not outbalanced by the gains in net number of products. Improved software designs are thus required to overtake this practical limitation.
\end{remark}

\subsection{Application: large data clustering}

\begin{figure}[t!]
    \centering
    \begin{tabular}{cc}
        \hspace*{-5mm}$K\!=\!\left(
        \vcenter{\hbox{
        \includegraphics[width=.3\linewidth,height=.3\linewidth]{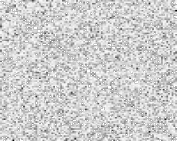}
        }}
        \right)$
        &
        \hspace*{-2mm}$K\!=\!\left(
        \vcenter{\hbox{
        \includegraphics[width=.3\linewidth,height=.3\linewidth]{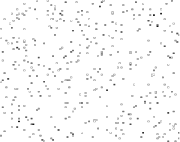}
        }}
        \right)$
        \vspace{1mm}
    \\
    \medskip
    \hspace*{-5mm}\includestandalone{fig_K_spectrum_S}
    &
    \hspace*{-5mm}\includestandalone{fig_K_spectrum_B}
    \\
    \hspace*{-5mm}\includestandalone{fig_eigvect_S}
    &
    \hspace*{-5mm}\includestandalone{fig_eigvect_B}
    \end{tabular}
     \vspace{-3mm}\\
    \caption{Two-way punctured matrices $K$ for {\bf (left)} $(\varepsilon_S,\varepsilon_B)=(.2,1)$ or {\bf (right)} $(\varepsilon_S,\varepsilon_B)=(1,.04)$, with $c_0=\frac12$, $n=4\,000$, $p=2\,000$, $b=0$. Clustering setting with $x_i\sim .4\mathcal N(\mu_1,I_p)+.6\mathcal N(\mu_2,I_p)$ for $[\mu_1^\T,\mu_2^\T]^\T\sim \mathcal N(0,\frac1p[\begin{smallmatrix} 20 & 12 \\ 12 & 30\end{smallmatrix}]\otimes I_p)$. {\bf (Top)} first $100\times 100$ absolute entries of $K$ (white for zero); {\bf (Middle)} spectrum of $K$, theoretical limit, and isolated eigenvalues; {\bf (Bottom)} second dominant eigenvector $\hat v_2$ of $K$ against theoretical average in red. \textbf{\textit{As confirmed by theory, although (top) $K$ is dense for $\varepsilon_B=1$ and sparse for $\varepsilon_B=.04$ ($96\%$ empty) and (middle) the spectra strikingly differ, (bottom) since $\varepsilon_S^2\varepsilon_Bc_0^{-1}$ is constant, the eigenvector alignment $|\hat v_2^\T v_2|^2$ is the same in both cases.}}  }
    \label{fig:clustering}
\end{figure}

As a telling application of our results, let us consider the spectral clustering setting described in Section~\ref{sec:PCA_spect_clust}.

\subsubsection{Synthetic data}

We first let $x_1,\ldots,x_n\in\mathbb R^p$ arise from a synthetic two-class Gaussian mixture with $n=4\,000$ and $p=2\,000$. Two puncturing approaches are compared: (i) reducing the cost of the inner products $x_i^\T x_j$ using a $5$-fold ($\varepsilon_S=.2$ while $\varepsilon_B=1$) random puncturing of the data vectors $x_i$, versus (ii) a $25$-fold puncturing of the matrix $\frac1pX^\T X$ ($\varepsilon_B=.04$ while $\varepsilon_S=1$). Figure~\ref{fig:clustering} depicts (for a setting detailed in caption) the matrices $K$, their spectra and second dominant eigenvector $\hat v_2$ ($\hat v_1$ is not discriminating in this setting, due to $P^\H P$ having a dominant all-ones eigenvector). The reported scenario is interesting in that we purposely took $\varepsilon_B\varepsilon_S^2c_0^{-1}$ constant in both cases; as such, while the matrices $K$ and their spectra dramatically differ, eigenvector $\hat v_2$ is essentially the ``same'' in both matrices. This first confirms the theory but most importantly defies the natural intuition that so different matrices cannot possibly give rise to the same eigenvector structure and quality.

\medskip

\begin{figure}[t!]
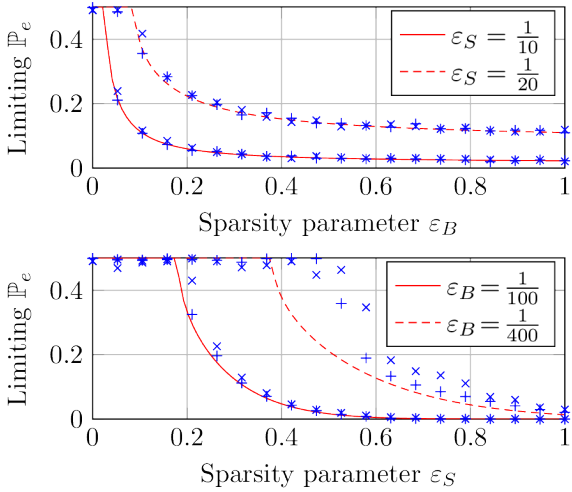

    \centering
    \begin{tabular}{c}
      \includestandalone{fig_perf_B}\\
      \includestandalone{fig_perf_S}
    \end{tabular}
    \vspace{-3mm}\\
    \caption{Limiting probability of error of spectral clustering of $\mathcal N(\pm\mu,I_p)$ with equal class sizes on $K$: as a function of
    $\varepsilon_B$ for fixed $\ell=\|\mu\|^2=50$ {\bf (top)}, and $\varepsilon_S$ for fixed $\ell=50$ {\bf (bottom)}. Simulations (single realization) in markers for $p=n=4\,000$ ($\color{blue}\times$) and $p=n=8\,000$ ($\color{blue}+$). \textit{\textbf{Very good fit between theory and practice for not too small $\varepsilon_S,\varepsilon_B$}}. }
    \label{fig:clustering_error}
\end{figure}

In the very symmetric setting of two classes of equal sizes ($n/2$ elements per class) and opposed statistical means (i.e., with $x_i\sim .5\mathcal N(\mu,I_p)+.5\mathcal N(-\mu,I_p)$), only one spike population eigenvalue is non-zero and $v=v_1$ is known: its normalized entries belong to $\{\pm\frac1{\sqrt n}\}$ (indeed, here $\mathcal M=\frac12\|\mu\|^2[\begin{smallmatrix} 1 & -1 \\ -1 & 1\end{smallmatrix}]$, the eigenvalues of which equal $\|\mu\|^2$ and $0$ with respective eigenvectors $[1,-1]$ and $[1,1]$). By symmetry, the random entries of the sample eigenvector $\hat v\equiv\hat v_1$ are asymptotically centered on $\pm \sqrt{\zeta/n}$ with variance asymptotically equal to $(1-\zeta)/n$ for $\zeta\equiv\zeta_1$ provided by Theorem~\ref{th:spikes} (with $\ell_1=\|\mu\|^2$). Related random matrix studies (e.g., \cite{kadavankandy2019asymptotic} for $\varepsilon_S=\varepsilon_B=1$) have shown that the fluctuations of the entries of $\hat v$ are asymptotically Gaussian and pairwise independent; this suffices to justify that the asymptotic classification error $\mathbb P_e$ incurred by spectral clustering is given by:
\begin{align*}
    \mathbb P_e = \frac1n \sum_{i=1}^n \delta_{\left\{ {\rm sign}([\hat v]_i[v]_i) < 0 \right\} } &\to Q\left( \sqrt{{\zeta}/{(1-\zeta)}} \right)
\end{align*}
almost surely, where $Q(t)=\frac1{\sqrt{2\pi}}\int_t^\infty e^{-u^2/2}du$ is the Gaussian tail function, and the (arbitrary) signs of $v,\hat v$ are chosen such that $0\leq P_e\leq \frac12$. Figure~\ref{fig:clustering_error} depicts the limiting error for various values of $(\varepsilon_S,\varepsilon_B,c_0,\ell)$. Despite $\varepsilon_B$ and $\varepsilon_S$ being particularly in this setting, the simulations show a strong fit between theory and practice, even for not so large values of $n$.

\begin{remark}[How large should $n,p$ be in practice?]
    It is well established in random matrix theory that limiting results can be obtained at speeds up to $O(1/\sqrt{pn})=O(1/n)$. We may in particular show here that $\mathbb P_e=Q(\sqrt{\zeta/(1-\zeta)})+O(1/n)$. As a consequence, our practical predictions are already accurate for quite small values of $n$.

    This being said, the $O(1/n)$ term hides constants, particularly depending on $\varepsilon_S,\varepsilon_B$ which cannot be taken too small. As a rule of thumb, $1/\varepsilon_S,1/\varepsilon_B$ must remain small compared to $p,n$.\footnote{If not, as discussed in the article concluding remarks, $K$ falls into a ``sparse regime'' no longer supported by the present random matrix analysis.} This last remark explains in passing the disrupted behavior of Figure~\ref{fig:clustering_error}-(bottom) for too small $\varepsilon_B$.
\end{remark}

\subsubsection{Resilience to real-world images}

\begin{figure}[tb]
     \centering
     \subfloat{\includegraphics[width=.19\linewidth]{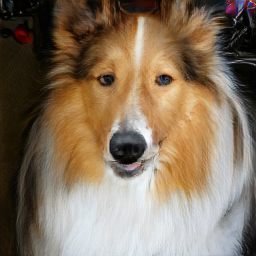}}\hspace{0.1mm}
     \subfloat{\includegraphics[width=.19\linewidth]{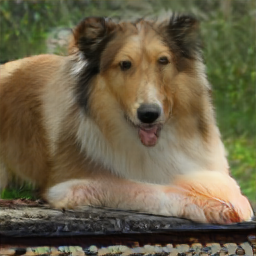}}\hspace{0.1mm}
     \subfloat{\includegraphics[width=.19\linewidth]{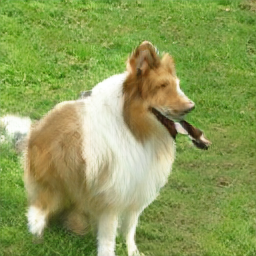}}\hspace{0.1mm}
     \subfloat{\includegraphics[width=.19\linewidth]{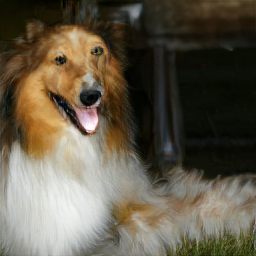}}\hspace{0.1mm}
     \subfloat{\includegraphics[width=.19\linewidth]{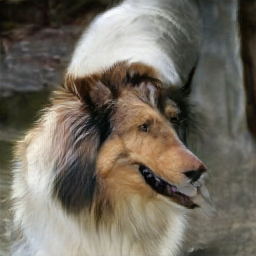}}\\
     \subfloat{\includegraphics[width=.19\linewidth]{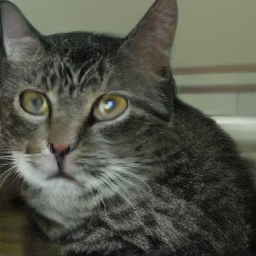}}\hspace{0.1mm}
     \subfloat{\includegraphics[width=.19\linewidth]{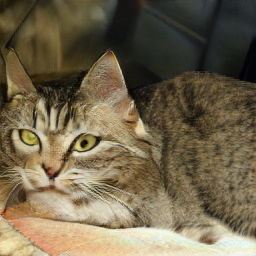}}\hspace{0.1mm}
     \subfloat{\includegraphics[width=.19\linewidth]{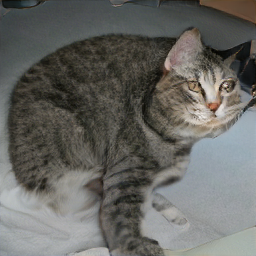}}\hspace{0.1mm}
     \subfloat{\includegraphics[width=.19\linewidth]{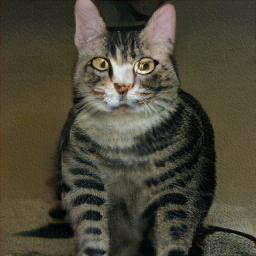}}\hspace{0.1mm}
     \subfloat{\includegraphics[width=.19\linewidth]{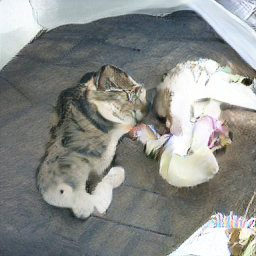}}
     \caption{Examples of BigGAN-generated images, `\texttt{collie}' dog instances ({\bf top row}),  `\texttt{tabby}' cat instances ({\bf bottom row}).}
     \label{fig:gan_example}
\end{figure}

\begin{figure}[tb]
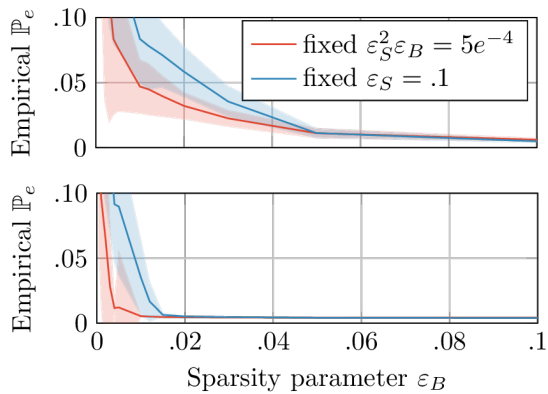

    \centering
    \begin{tabular}{c}
        \includestandalone{fig_perf_gan_small_n}\\
        \, \includestandalone{fig_perf_gan_large_n}
    \end{tabular}
    \vspace{-3mm}\\
    \caption{Empirical classification errors for $2$-class (balanced) BigGAN-generated images (`\texttt{tabby}' vs `\texttt{collie}'), with $n=2\,500$ ({\bf top}) and $n=10\,000$ ({\bf bottom}). \textit{\textbf{Theoretically predicted ``plateau''-behavior observed for all $\varepsilon_B$ not too small}}. }
    \label{fig:spectrum_GAN}
\end{figure}

\begin{figure}[tb]
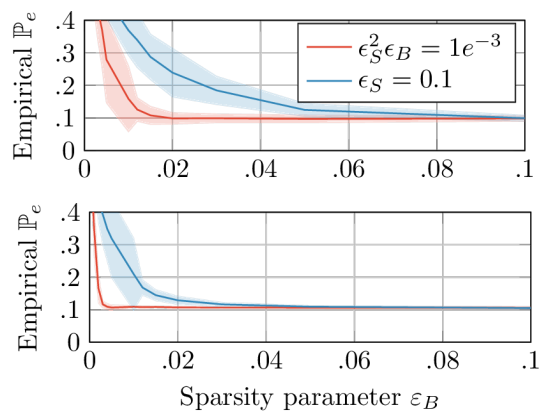

    \centering
    \begin{tabular}{c}
        \includestandalone{fig_perf_mnist_n_small}\\
        \, \includestandalone{fig_perf_mnist_n_large}
    \end{tabular}
    \vspace{-3mm}\\
    \caption{Empirical classification errors for $2$-class (balanced) MNIST-fashion images (`\texttt{trouser}' vs `\texttt{pullover}'), with $n=512$ ({\bf top}) and $n=2048$ ({\bf bottom}). \textit{\textbf{Similar ``plateaus'' as predicted by the theory and observed in Figure~\ref{fig:spectrum_GAN}}}.
    }
    \label{fig:spectrum_MNIST}
\end{figure}

To practically confirm our theoretical findings, we next apply the two-way puncturing kernel to vectors $x_i$ arising from a two-class mixture (`\texttt{tabby}' cats versus `\texttt{collie}' dogs; see Figure~\ref{fig:gan_example}) of the (globally centered and scaled) $p=4\,096$-VGG features of randomly BigGAN-generated images \cite{brock2018large}. The results are for varying $\varepsilon_B$ and either fixed $\varepsilon_S$ or $\varepsilon_S$ set such that $\varepsilon_S^2\varepsilon_B=5\cdot 10^{-4}$. The simulation depicted in Figure~\ref{fig:spectrum_GAN} corroborates the presence of a performance ``plateau'' and a significant reduction of the transition value of $\varepsilon_B$ (from $.05$ to $.015$) when $n$ (and thus $1/c_0$) increases fourfold. This supports the theoretical performance of the central display in Figure~\ref{fig:clustering_error}. Maintaining $\varepsilon_S^2\varepsilon_B$ constant pushes this plateau further down to smaller values of $\varepsilon_B$ until the method breaks. The same conclusion can be drawn on non-pretreated $p=784$-dimensional  real word images from the Fashion-MNIST dataset, as shown in Figure~\ref{fig:spectrum_MNIST}.

\begin{figure}[htb!]
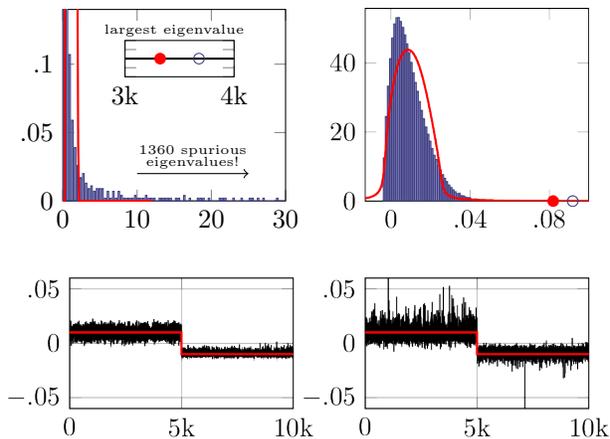

\centering
\hspace*{-3mm}\begin{tabular}{cc}
    \subfloat{\includestandalone{fig_gan_spectrum_full}} &\hspace*{-8mm}\subfloat{\includestandalone{fig_gan_spectrum}}
    \\ 
    \subfloat{\includestandalone{fig_gan_vep_full}} &\hspace*{-8mm}\subfloat{\includestandalone{fig_gan_vep}}
\end{tabular}
     \caption{Sample vs limiting spectra and dominant eigenvector of $K$ for 2-class GAN images (\texttt{tabby} vs \texttt{collie}); {\bf(left)} $\varepsilon_S=\varepsilon_B=1$ (error rate: $\mathbb P_e=.004$); {\bf (right)} $\varepsilon_S=0.01$, $\varepsilon_B=0.2$ ($\mathbb P_e=.011$). \textit{\textbf{Surprisingly good fit between sample and predicted isolated eigenvalue/eigenvector in all cases; as for spectral measure, significant prediction improvement as $\varepsilon_S,\varepsilon_B\to 0$}}.}
     \label{fig:gan_hist}
\end{figure}

More interestingly, as shown in Figure~\ref{fig:gan_hist}, while for $\varepsilon_B=\varepsilon_S=1$ the eigenvalues of $K$ for the GAN images spread far from the theoretical Mar\u{c}enko-Pastur limit,\footnote{This may at first be thought to follow from strong feature covariance (thus not close to $I_p$), but it turns out that in-sample correlation is even stronger as the VGG-features of the produced GAN images appear to have a very low variability.} for $\varepsilon_B,\varepsilon_S\ll 1$, the empirical spectrum is very close to the predicted (uncorrelated vector) limit: this strongly suggests that intensive puncturing has the effect to ``decorrelate'' data. This remark has the powerful advantage to improve the theoretical tractability of these preprocessed data. More surprisingly, for both small or large $\varepsilon_B,\varepsilon_S$, despite the general spectrum mismatch, the anticipated dominant eigenvalue position and eigenvector behavior are extremely good, making it still possible to predict clustering performance with good accuracy. The same conclusions apply to the Fashion-MNIST dataset (see figures in the gitlab repository).


\section{Concluding remarks}
\label{sec:conclusion}

A fundamental conclusion of the article, confirmed on practical data, is that drastic computation and storage reduction can be theoretically achieved while virtually incurring no loss in PCA or spectral clustering. This follows from the peculiar behavior of (doubly) punctured kernel and sample covariance matrices $K$. As shown in an enlarging spectrum of articles, the large dimensional behavior of $Q$ has immediate further implications to the performance behavior of many machine learning algorithms, ranging from support vector machines \cite{kammoun2020precise,huang2017asymptotic} to semi-supervised graph inference \cite{mai2018random}, transfer and multi-task learning \cite{tiomoko2020large}, random feature maps \cite{liao2018spectrum,pennington2019nonlinear}, or neural network dynamics \cite{liao2018dynamics,advani2020high}, to cite a few. As such, the article, rather than providing a ready-to-use method for fast unsupervised learning, really lays the theoretical ground to a systematic cost and storage reduction approach to a host of learning algorithms.

On the downside though, following up on Remark~\ref{rem:cache}, the effective software libraries for sparse matrix operations (which heavily rely on block-sparsity) are far from optimal when compared to efficient dense matrix operations, and thus demand a profound treatment to ensure that our claimed computational cost improvements are truly met in practice. This is not a negligible aspect of the puncturing framework which we shall investigate in greater depth in the future.

Another critical aspect lies in the request that $\varepsilon_B,\varepsilon_S=O(1)$ with respect to $p,n$, thereby \emph{not allowing for truly sparse} $K$. For more severe puncturing, random matrix theory fails to provide accurate predictions and, worse, the optimal phase transition threshold is no longer met by clustering from $K$ but from more elaborate matrices (such as proposed by statistical physicists \cite{krzakala2013spectral,dall2019revisiting}). Pushing towards sparser models therefore demands a dramatic change of theoretical standpoint.

\section*{Acknowledgment}

Couillet's work is supported by the ANR-MIAI Large-DATA chair at University Grenoble-Alpes (ANR-19-P3IA-0003), and the HUAWEI-GIPSA LarDist project.

\bibliography{references}

\begin{thebibliography}{26}
\providecommand{\natexlab}[1]{#1}
\providecommand{\url}[1]{\texttt{#1}}
\expandafter\ifx\csname urlstyle\endcsname\relax
  \providecommand{\doi}[1]{doi: #1}\else
  \providecommand{\doi}{doi: \begingroup \urlstyle{rm}\Url}\fi

\bibitem[Advani et~al.(2020)Advani, Saxe, and Sompolinsky]{advani2020high}
Advani, M.~S., Saxe, A.~M., and Sompolinsky, H.
\newblock High-dimensional dynamics of generalization error in neural networks.
\newblock \emph{Neural Networks}, 132:\penalty0 428--446, 2020.

\bibitem[Bottou(1991)]{bottou1991stochastic}
Bottou, L.
\newblock Stochastic gradient learning in neural networks.
\newblock \emph{Proceedings of Neuro-N{\i}mes}, 91\penalty0 (8):\penalty0 12,
  1991.

\bibitem[Brock et~al.(2018)Brock, Donahue, and Simonyan]{brock2018large}
Brock, A., Donahue, J., and Simonyan, K.
\newblock Large scale gan training for high fidelity natural image synthesis.
\newblock \emph{arXiv preprint arXiv:1809.11096}, 2018.

\bibitem[Cai et~al.(2013)Cai, Ma, Wu, et~al.]{cai2013sparse}
Cai, T.~T., Ma, Z., Wu, Y., et~al.
\newblock Sparse pca: Optimal rates and adaptive estimation.
\newblock \emph{The Annals of Statistics}, 41\penalty0 (6):\penalty0
  3074--3110, 2013.

\bibitem[Dall'Amico et~al.(2019)Dall'Amico, Couillet, and
  Tremblay]{dall2019revisiting}
Dall'Amico, L., Couillet, R., and Tremblay, N.
\newblock Revisiting the bethe-hessian: improved community detection in sparse
  heterogeneous graphs.
\newblock In \emph{33rd Conference on Neural Information Processing Systems
  (NeurIPS 2019), Vancouver, Canada}, pp.\  4039--4049, 2019.

\bibitem[Deshpande \& Montanari(2014)Deshpande and
  Montanari]{deshpande2014information}
Deshpande, Y. and Montanari, A.
\newblock Information-theoretically optimal sparse pca.
\newblock In \emph{2014 IEEE International Symposium on Information Theory},
  pp.\  2197--2201. IEEE, 2014.

\bibitem[Engel et~al.(2004)Engel, Mannor, and Meir]{engel2004kernel}
Engel, Y., Mannor, S., and Meir, R.
\newblock The kernel recursive least-squares algorithm.
\newblock \emph{IEEE Transactions on signal processing}, 52\penalty0
  (8):\penalty0 2275--2285, 2004.

\bibitem[Freund et~al.(2007)Freund, Dasgupta, Kabra, and
  Verma]{freund2007learning}
Freund, Y., Dasgupta, S., Kabra, M., and Verma, N.
\newblock Learning the structure of manifolds using random projections.
\newblock In \emph{NIPS}, volume~7, pp.\ ~59. Citeseer, 2007.

\bibitem[Huang(2017)]{huang2017asymptotic}
Huang, H.
\newblock Asymptotic behavior of support vector machine for spiked population
  model.
\newblock \emph{The Journal of Machine Learning Research}, 18\penalty0
  (1):\penalty0 1472--1492, 2017.

\bibitem[Johnstone \& Lu(2009)Johnstone and Lu]{johnstone2009sparse}
Johnstone, I.~M. and Lu, A.~Y.
\newblock Sparse principal components analysis.
\newblock \emph{arXiv preprint arXiv:0901.4392}, 2009.

\bibitem[Kadavankandy \& Couillet(2019)Kadavankandy and
  Couillet]{kadavankandy2019asymptotic}
Kadavankandy, A. and Couillet, R.
\newblock Asymptotic gaussian fluctuations of spectral clustering eigenvectors.
\newblock In \emph{2019 IEEE 8th International Workshop on Computational
  Advances in Multi-Sensor Adaptive Processing (CAMSAP)}, pp.\  694--698. IEEE,
  2019.

\bibitem[Kammoun \& Alouini(2020)Kammoun and Alouini]{kammoun2020precise}
Kammoun, A. and Alouini, M.-S.
\newblock On the precise error analysis of support vector machines.
\newblock \emph{arXiv preprint arXiv:2003.12972}, 2020.

\bibitem[Keriven et~al.(2018)Keriven, Bourrier, Gribonval, and
  P{\'e}rez]{keriven2018sketching}
Keriven, N., Bourrier, A., Gribonval, R., and P{\'e}rez, P.
\newblock Sketching for large-scale learning of mixture models.
\newblock \emph{Information and Inference: A Journal of the IMA}, 7\penalty0
  (3):\penalty0 447--508, 2018.

\bibitem[Krzakala et~al.(2013)Krzakala, Moore, Mossel, Neeman, Sly,
  Zdeborov{\'a}, and Zhang]{krzakala2013spectral}
Krzakala, F., Moore, C., Mossel, E., Neeman, J., Sly, A., Zdeborov{\'a}, L.,
  and Zhang, P.
\newblock Spectral redemption in clustering sparse networks.
\newblock \emph{Proceedings of the National Academy of Sciences}, 110\penalty0
  (52):\penalty0 20935--20940, 2013.

\bibitem[Liao \& Couillet(2018{\natexlab{a}})Liao and
  Couillet]{liao2018dynamics}
Liao, Z. and Couillet, R.
\newblock The dynamics of learning: A random matrix approach.
\newblock In \emph{International Conference on Machine Learning}, pp.\
  3072--3081. PMLR, 2018{\natexlab{a}}.

\bibitem[Liao \& Couillet(2018{\natexlab{b}})Liao and
  Couillet]{liao2018spectrum}
Liao, Z. and Couillet, R.
\newblock On the spectrum of random features maps of high dimensional data.
\newblock In \emph{International Conference on Machine Learning}, pp.\
  3063--3071. PMLR, 2018{\natexlab{b}}.

\bibitem[Mai \& Couillet(2018)Mai and Couillet]{mai2018random}
Mai, X. and Couillet, R.
\newblock A random matrix analysis and improvement of semi-supervised learning
  for large dimensional data.
\newblock \emph{The Journal of Machine Learning Research}, 19\penalty0
  (1):\penalty0 3074--3100, 2018.

\bibitem[Mar{\v{c}}enko \& Pastur(1967)Mar{\v{c}}enko and
  Pastur]{marvcenko1967distribution}
Mar{\v{c}}enko, V.~A. and Pastur, L.~A.
\newblock Distribution of eigenvalues for some sets of random matrices.
\newblock \emph{Mathematics of the USSR-Sbornik}, 1\penalty0 (4):\penalty0 457,
  1967.

\bibitem[Murtagh \& Contreras(2012)Murtagh and
  Contreras]{murtagh2012algorithms}
Murtagh, F. and Contreras, P.
\newblock Algorithms for hierarchical clustering: an overview.
\newblock \emph{Wiley Interdisciplinary Reviews: Data Mining and Knowledge
  Discovery}, 2\penalty0 (1):\penalty0 86--97, 2012.

\bibitem[Pastur \& Shcherbina(2011)Pastur and Shcherbina]{pastur2011eigenvalue}
Pastur, L.~A. and Shcherbina, M.
\newblock \emph{Eigenvalue distribution of large random matrices}.
\newblock Number 171. American Mathematical Soc., 2011.

\bibitem[Pennington \& Worah(2019)Pennington and
  Worah]{pennington2019nonlinear}
Pennington, J. and Worah, P.
\newblock Nonlinear random matrix theory for deep learning.
\newblock \emph{Journal of Statistical Mechanics: Theory and Experiment},
  2019\penalty0 (12):\penalty0 124005, 2019.

\bibitem[Tiomoko et~al.(2020)Tiomoko, Couillet, and Tiomoko]{tiomoko2020large}
Tiomoko, M., Couillet, R., and Tiomoko, H.
\newblock Large dimensional analysis and improvement of multi task learning.
\newblock \emph{arXiv preprint arXiv:2009.01591}, 2020.

\bibitem[Von~Luxburg(2007)]{von2007tutorial}
Von~Luxburg, U.
\newblock A tutorial on spectral clustering.
\newblock \emph{Statistics and computing}, 17\penalty0 (4):\penalty0 395--416,
  2007.

\bibitem[Wigner(1958)]{wigner1958distribution}
Wigner, E.~P.
\newblock On the distribution of the roots of certain symmetric matrices.
\newblock \emph{Annals of Mathematics}, pp.\  325--327, 1958.

\bibitem[Zarrouk et~al.(2020)Zarrouk, Couillet, Chatelain, and
  Le~Bihan]{zarrouk2020performance}
Zarrouk, T., Couillet, R., Chatelain, F., and Le~Bihan, N.
\newblock Performance-complexity trade-off in large dimensional statistics.
\newblock In \emph{2020 IEEE 30th International Workshop on Machine Learning
  for Signal Processing (MLSP)}, pp.\  1--6. IEEE, 2020.

\bibitem[Zhong et~al.(2020)Zhong, Su, and Fan]{zhong2020empirical}
Zhong, X., Su, C., and Fan, Z.
\newblock Empirical bayes pca in high dimensions.
\newblock \emph{arXiv preprint arXiv:2012.11676}, 2020.

\end{thebibliography}
\bibliographystyle{icml2020}

\end{document}


\title{Two-way kernel matrix puncturing:\\ towards resource-efficient\\ PCA and spectral clustering \\ \ \\ --- Supplementary Material ---}

\maketitle

\begin{abstract}
    This supplementary material provides the proofs of the main theorems of the core article.
\end{abstract}

\section{Reminder of the main setting}

For convenience, we first recall our main setting and assumptions. The central object of interest is the matrix
\begin{align}
    \label{eq:K}
    K &= \left\{\frac1p(X\odot S)^\H(X\odot S)\right\} \odot B \in\mathbb C^{n\times n}
\end{align}
under the large dimensional $n,p$ regime. Here, $X$, $S$ and $B$ satisfy the following assumptions.

\begin{assumption}[Data model]
\label{ass:model}
\begin{align*}
    X &= Z + P
\end{align*}
where the $Z_{ij}\sim \mathcal{CN}(0,1)$ are independent, and where $P\in\mathbb C^{p\times n}$ is a rank-$k$ matrix for some $k$.
\end{assumption}

\begin{assumption}[Large $p,n$ asymptotics]
\label{ass:asymptotics}
As $n\to\infty$,
\begin{align*}
    p/n &\to c_0\in(0,\infty)
\end{align*}
and there exists a decomposition $P=LV^\H$ of $P$ with $V\in\mathbb C^{n\times k}$ isometric (i.e., $V^\H V=I_k$) and
\begin{align*}
    \frac1pL^\H L &\to \mathcal L
\end{align*}
for some deterministic matrix $\mathcal L \in\mathbb C^{k\times k}$. In particular, the eigenvalues of $\mathcal L$ are the limiting $k$ non-trivial eigenvalues of $\frac1pP^\H P$. Besides,
\begin{align*}
    \limsup_n\max_{ \substack{1\leq i\leq n \\ 1\leq j\leq k}}\{\sqrt nV_{ij}^2\}=0.
\end{align*}
\end{assumption}

\section{The theorems}

The spectral characterization of $K$ is made through the study of its resolvent matrix
\begin{align*}
    Q(z) &= (K-zI_n)^{-1}.
\end{align*}

The results are then as follows.

\begin{theorem}[Deterministic equivalent for $Q$]
    \label{th:barQ}
    Under Assumptions~\ref{ass:model}--\ref{ass:asymptotics}, let $z\in\mathbb C$ be away from the limsup of the union of supports of $\nu_1,\nu_2,\ldots$. Then, as $n\to\infty$,
    \begin{align*}
        Q(z) &\leftrightarrow m(z) \left[I_n + \frac{c_0^{-1}\varepsilon_S^2\varepsilon_B m(z)}{1+\varepsilon_B\varepsilon_Sc_0^{-1}m(z)}V\mathcal LV^\H\right]^{-1}
    \end{align*}
    where $m(\cdot)$ is the unique Stieltjes transform solution to\footnote{
    We also recall that the notation $A\leftrightarrow B$ stands for the fact that, for any linear functional $u:\mathbb C^{n\times n}\to \mathbb R$ of bounded infinity norm, $u(A-B)\to 0$ almost surely as $n\to\infty$.
    }
    \begin{align*}
    z = \varepsilon_Sb - \frac1{m(z)} - c_0^{-1}\varepsilon_B\varepsilon_S^2m(z)+\frac{c_0^{-2}\varepsilon_B^3\varepsilon_S^3m(z)^2}{1+c_0^{-1}\varepsilon_B\varepsilon_S m(z)}.
\end{align*}
\end{theorem}

This theorem is in fact sufficiently exhaustive to characterize both the \emph{macroscopic} spectrum of $K$ (its limiting spectral measure) as well as the \emph{microscopic} behavior of its dominant isolated eigenvalues and associated eigenvectors. The next result, which we name theorem in compliance with the core article, is in effect an (important) corollary of Theorem~\ref{th:barQ}.

\begin{theorem}[Phase transition, isolated eigenvalues and eigenvectors]
    \label{th:spikes}
    Define the functions
    \begin{align*}
        F(t)&=t^4+\frac{2}{\varepsilon_S}t^3+\frac{1}{\varepsilon_S^2}\left( 1 - \frac{c_0}{\varepsilon_B}\right)t^2 - \frac{ 2c_0}{\varepsilon_S^3}t-\frac{c_0}{\varepsilon_S^4} \\
        G(t)&= \varepsilon_Sb+c_0^{-1}\varepsilon_B\varepsilon_S(1+\varepsilon_St)+\frac{\varepsilon_S}{1+\varepsilon_St}+\frac{\varepsilon_B}{t(1+\varepsilon_St)}
    \end{align*}
    and $\Gamma\in\mathbb R$ be the largest real solution to $F(\Gamma)=0$.
    Further denote $\ell_1>\ldots>\ell_{\bar k}$ the $\bar k\leq k$ \emph{distinct} eigenvalues of $\mathcal L$ of respective multiplicities $L_1,\ldots,L_{\bar k}$, and $\Pi_1,\ldots,\Pi_{\bar k}\in\mathbb R^{k\times k}$ the projectors on their respective associated eigenspaces. Similarly denote $(\lambda_1,\hat v_1),\ldots,(\lambda_n,\hat v_n)$ the eigenvalue-eigenvector pairs of $K$ in descending order and gather the first $k$ eigenvectors under the isometric matrices $\hat{\mathcal V}_1=[\hat v_1,\ldots,\hat v_{L_1}]$ up to $\hat{\mathcal V}_{\bar k}=[\hat v_{k-L_{\bar k}+1},\ldots,\hat v_k]$.

    Then, for $i\in\{1,\ldots,\bar k\}$ and for all $j\in\{L_1+\ldots+L_{i-1}+1,\ldots,L_1+\ldots+L_i\}$,
    \begin{align*}
        \lambda_j &\to \rho_i \equiv \left\{
        \begin{array}{ll}
             F(\ell_i) &,~\ell_i>\Gamma  \\
             F(\Gamma) &,~\ell_i\leq \Gamma
        \end{array}\right.
        \end{align*}
        almost surely, and
        \begin{align*}
        \hat{\mathcal V}_i\hat{\mathcal V}_i^\H \leftrightarrow \zeta_i V\Pi_i V^\H,~\textmd{where}~\zeta_i = \left\{
        \begin{array}{ll} \frac{F(\ell_i)\varepsilon_S^3}{\ell_i(1+\varepsilon_S\ell_i)^3} &,~\ell_i>\Gamma \\
        0&,~\ell_i\leq\Gamma
        \end{array}
        \right.
    \end{align*}
    with the notation `$\leftrightarrow$' introduced in Theorem~\ref{th:barQ}. In particular, if the $\ell_i$'s have unit multiplicities with associated population eigenvectors $v_i$, then
    \begin{align*}
        |v_i^\H \hat v_i|^2 \to \zeta_i,\quad i=1,\ldots,k.
    \end{align*}
\end{theorem}

\section{Elements of proof}

\subsection{Rationale}

The proof relies on the \emph{Gaussian tools} for random matrices popularized in \cite{} and consisting in exploiting Stein's lemma
\begin{align*}
    \mathbb{E}[z\phi(z)]=\mathbb{E}\left[\frac{\partial}{\partial \bar z} \phi(z)\right]
\end{align*}
for standard complex (or real) Gaussian random variables $z\sim \mathcal{CN}(0,1)$ along with the Nash-Poincar\'e inequality
\begin{align*}
    {\rm Var}[f(z)] \leq \sum_{i=1}^n \left( \mathbb{E}\left[ \left|\frac{\partial}{\partial z_i} f(z) \right|^2 \right]  + \mathbb{E}\left[ \left|\frac{\partial}{\partial \bar z_i} f(z) \right|^2 \right] \right)
\end{align*}
for standard multivariate complex Gaussian $z\sim \mathcal{CN}(0,I_n)$.

\medskip

As we shall see, Stein's lemma is used to ``\emph{unfold}'' the a priori quite involved form of the expected value $\mathbb E[Q_{ij}]$ of the entries of the resolvent matrix $Q$ of $K$. The Nash-Poincar\'e inequality is then subsequently used to control that the variance of $Q_{ij}$ vanishes at a proper rate.

\subsection{Proof of Theorem~1}

In order to be in a position to apply Stein's lemma, we first exploit the straightforward \emph{resolvent identity}: $KQ-zQ=I_n$, so to obtain
\begin{align*}
    \EE [Q_{ij}] &= -\frac1z\delta_{ij} + \frac1z \EE[[KQ]_{ij}].
\end{align*}

By expanding $X=Z+P$, with $P$ decomposed as $P=LV^\H$ ($L\in\mathbb C^{p\times k}$ and $V\in\mathbb C^{n\times k}$), we have to consider four terms in the expansion of $\EE[[KQ]_{ij}]$.

\subsection*{Term 1: involving $(Z^\H,Z)$}
Anticipating coming results, instead of evaluating $\EE[[KQ]_{ij}]$ directly, we rather evaluate a modified version in which matrix $B$ is replaced by a deterministic matrix $A$ with bounded operator norm and bounded entries: using Stein's lemma, we have
\begin{align}
    &\EE\left[ \left[\left(\left[\frac1p(Z\odot S)^\H(Z\odot S)\right]\odot A\right) Q\right]_{ij} \right] \nonumber \\
    &= \frac1p\sum_{l=1}^p\sum_{m=1}^n S_{li} S_{lm} A_{im} \EE\left[ \bar Z_{li} Z_{lm} Q_{mj} \right] \nonumber \\
    &=\frac1p\sum_{l=1}^p\sum_{m=1}^n S_{li} S_{lm} A_{im} \left(\EE\left[\delta_{im} Q_{mj} + Z_{lm} \frac{\partial Q_{mj}}{\partial Z_{li}} \right] \right).
    \label{eq:term1}
\end{align}
Using $\frac{\partial Q}{\partial Z_{ab}} = -Q \frac{\partial Q}{\partial Z_{ab}} Q$, it then comes
\begin{align*}
    \frac{\partial Q_{cd}}{\partial Z_{ab}} &= -\frac{1}{p}
    \sum_{l,l'=1}^n Q_{il} \left[\left[ (X \odot S)' (E_{ab} \odot S) \right] \odot B \right]_{ll'} Q_{l'd},
\end{align*}
for $E_{ab}$ the matrix with all zero entries but at coordinate $(a,b)$ where the entry equals $1$. We further have that
\begin{align*}
    \left[\left[ (X \odot S)^\H (E_{ab} \odot S) \right] \odot B \right]_{ll'}= \sum_{o=1}^p \bar X_{ol} S_{ol} \delta_{oa}  S_{ab} B_{ll'}= \bar X_{al} S_{al} S_{ab} B_{ll'} \delta_{l'b}
\end{align*}
so that
\begin{align*}
    \frac{\partial Q_{cd}}{\partial Z_{ab}} &= -\frac{1}{p} \left[ Q D_{B_{\cdot, b}} (X \odot S)^\H \right]_{ca} S_{ab} Q_{bd}.
\end{align*}
We then obtain for $T_1(A,S) \equiv \EE\left[ \left[\left(\left[\frac1p(Z\odot S)^\H(Z\odot S)\right]\odot A\right) Q\right]_{ij} \right]$ in \eqref{eq:term1}:
\begin{align*}
     T_1(A,S) &=
    \frac1p \sum_{l=1}^p\sum_{m=1}^n S_{li} S_{lm} A_{im} \EE\left[\delta_{im} Q_{mj}
    -\frac1p Z_{lm} \left[Q  D_{B_{\cdot, i}} (X \odot S)^\H \right]_{ml} S_{li} Q_{ij}\right]\\
    &= \EE\left[ \frac1p\left[S^\H S\right]_{ii} A_{ii} Q_{ij}\right] - \frac1{p^2} \sum_{l=1}^p \left[ Z D_{ S_{l,\cdot} } D_{A_{i,\cdot}} Q D_{ B_{\cdot,i} } (X \odot S)^\H D_{ S_{\cdot,i} } D_{S_{\cdot,i}} \right]_{ll}  Q_{ij}
\end{align*}
where $D_{x}$ denotes the diagonal matrix with elements the entries of vector $x$.

\subsection*{Term 2: $(Z',P)$}

We obtain for $T_2(A,P) \equiv \EE\left[ \left[\left(\left[\frac1p(Z\odot S)^\H(P\odot S)\right]\odot A\right) Q\right]_{ij} \right]$:
\begin{align*}
     T_2(A,R) &= \sum_{l=1}^p \sum_{m=1}^n \frac1p \EE \left[
     \bar Z_{li} S_{li} P_{lm} S_{lm} A_{im} Q_{mj}\right],\\
     &= \frac1p  \sum_{l=1}^p \sum_{m=1}^n
      S_{li} P_{lm} S_{lm} A_{im} \EE \left[ \frac{\partial Q_{mj}}{\partial z_{li}}\right],\\
     &=- \frac{1}{p^2}  \sum_{l=1}^p \sum_{m=1}^n
      S_{li} P_{lm} S_{lm} A_{im} \EE \left[ \left(Q D_{B_{\cdot,i}} (X\odot S)^\H\right)_{ml} S_{li} Q_{ij} \right]\\
     &=-  \frac{1}{p^2}   \sum_{l=1}^p \EE \left[ \left[ T D_{S_{l,\cdot} } D_{ A_{i,\cdot}} Q D_{B_{\cdot,i}} (X\odot S)^\H  D_{S_{\cdot,i} } \right]_{ll} Q_{ij}
     \right]
\end{align*}
where we used in particular the fact that $D_{S_{\cdot,i} }^2=D_{S_{\cdot,i} }$.

\subsection*{Summation of $T_1(A,S)$ and $T_2(A,S)$:} Summing the two previous terms, we get
\begin{align*}
    T_1(A,S)+T_2(A,S) &=
     \EE\left[ \frac1p\left[S^\H S\right]_{ii} A_{ii} Q_{ij}\right] \nonumber \\
     &- \frac1{p^2} \sum_{l=1}^p\EE\left[ \left[ X D_{ S_{l,\cdot} } D_{ A_{i,\cdot}   } Q
    D_{B_{\cdot,i} } (X \odot S)^\H D_{ S_{\cdot,i} } \right]_{ll}  Q_{ij}\right]
\end{align*}
Due to the presence of the term $S_{l,\cdot}$ inside the matrix evaluated at position $(l,l)$, the summation over $l$ cannot be ``turned into a trace'', as conventionally done to prove e.g., the Mar\u{c}enko-Pastur theorem \cite{} (when $S_{ij}=1$ and $B_{ij}=1$ for all $i,j$). We therefore need to proceed otherwise by writing
\begin{align*}
    &\frac1{p^2} \sum_{l=1}^p \left[ X D_{ S_{l,\cdot}} D_{ A_{i,\cdot}   } Q D_{B_{\cdot,i} } (X \odot S)^\H D_{ S_{\cdot,i} } \right]_{ll} \\
    &= \frac1{p^2} \sum_{l=1}^p X_{l,\cdot} D_{ S_{l,\cdot} } D_{ A_{i,\cdot}   } Q D_{B_{\cdot,i} } D_{S_{l,\cdot}} X_{l,\cdot}^\H S_{l,i}.
\end{align*}
To evaluate the quadratic forms, we must ``break'' the dependence between $X_{l,\cdot}$ and $Q$. To this end, note that
\begin{align*}
    Q &= \left( \frac1p\sum_{i=1}^p \left\{\left[D_{S_{i,\cdot}} X_{i,\cdot}^\H X_{i,\cdot}D_{S_{i,\cdot}}\right]\odot B\right\} - z I_n\right)^{-1}
\end{align*}
so that, applying Woodbury's identity,
\begin{align*}
    Q &= Q_{-l}- \frac1p Q_{-l} \left\{\left[D_{S_{l,\cdot}} X_{i,\cdot}^\H X_{i,\cdot}D_{S_{l,\cdot}}\right]\odot B \right\}\left(I_n + \frac1pQ_{-l} \left\{\left[D_{S_{l,\cdot}} X_{i,\cdot}^\H X_{i,\cdot}D_{S_{l,\cdot}}\right]\odot B\right\} \right)^{-1}Q_{-l}.
\end{align*}

Plugged into the quadratic form over $X_{l,\cdot}$, this gives:
\begin{align*}
    &\frac1pX_{l,\cdot} D_{ S_{l,\cdot} } D_{ A_{i,\cdot}   } Q D_{B_{\cdot,i} } D_{S_{l,\cdot}} X_{l,\cdot}^\H \\
    &=\frac1p X_{l,\cdot} D_{ S_{l,\cdot} } D_{ A_{i,\cdot}   } Q_{-l} D_{B_{\cdot,i} } D_{S_{l,\cdot}} X_{l,\cdot}^\H \\
    &-\frac1{p^2} X_{l,\cdot} D_{ S_{l,\cdot} } D_{ A_{i,\cdot}   }
    Q_{-l} \left\{\left[D_{S_{l,\cdot}} X_{l,\cdot}^\H X_{l,\cdot}D_{S_{l,\cdot}}\right]\odot B \right\}\left(I_n + \frac1pQ_{-l} \left\{\left[D_{S_{l,\cdot}} X_{l,\cdot}^\H X_{l,\cdot}D_{S_{l,\cdot}}\right]\odot B\right\} \right)^{-1}\\
    &\times Q_{-l} D_{B_{\cdot,i} } D_{S_{l,\cdot}} X_{l,\cdot}^\H.
\end{align*}

Recalling that $X=Z+LV^\H$ (so that $X_{l,\cdot}=Z_{l,\cdot}+L_{l,\cdot}V^\H$), we first find that, averaging over $Z_{l,\cdot}$,
\begin{align*}
    &\frac1p\EE \left[ X_{l,\cdot} D_{ S_{l,\cdot} } D_{A_{i,\cdot}   } Q_{-l} D_{B_{\cdot,i} } D_{S_{l,\cdot}} X_{l,\cdot}^\H\right] \\
    &=\frac1p \EE \left[\tr D_{ S_{l,\cdot}  } D_{ A_{i,\cdot}   } Q_{-l} D_{B_{\cdot,i} } D_{S_{l,\cdot}}\right] + \frac1p \EE\left[ L_{l,\cdot}V^\H D_{ S_{l,\cdot} } D_{ A_{i,\cdot}   } Q_{-l} D_{B_{\cdot,i} } D_{S_{l,\cdot}}VL_{l,\cdot}^\H\right].
\end{align*}
A further application of the Nash-Poincar\'e inequality then shows that the variance of $X_{l,\cdot} D_{ S_{l,\cdot} } D_{A_{i,\cdot}   } Q_{-l} D_{B_{\cdot,i} } D_{S_{l,\cdot}} X_{l,\cdot}^\H$ vanishes as $O(1/p)$ while ${\rm Var}[Q_{ij}]=O(1)$, so that the above result extends into
\begin{align*}
   &\frac1p\EE \left[ X_{l,\cdot} D_{ S_{l,\cdot} } D_{A_{i,\cdot}   } Q_{-l} D_{B_{\cdot,i} } D_{S_{l,\cdot}} X_{l,\cdot}^\H Q_{ij}\right] \nonumber \\
   &= \frac1p \EE \left[\tr D_{ S_{l,\cdot}  } D_{ A_{i,\cdot}   } Q_{-l} D_{B_{\cdot,i} } D_{S_{l,\cdot}}\right]\EE[Q_{ij}] \nonumber \\
   &+ \frac1p \EE\left[ L_{l,\cdot}V^\H D_{ S_{l,\cdot} } D_{ A_{i,\cdot}   } Q_{-l} D_{B_{\cdot,i} } D_{S_{l,\cdot}}VL_{l,\cdot}^\H\right]\EE[Q_{ij}] + O(p^{-\frac12}).
\end{align*}

Now observe, for the second right-hand side term, that, by Cauchy-Schwarz's inequality and after summation over $l$,
\begin{align*}
    &\left(\frac1{p^2} \sum_{l=1}^p \EE\left[ L_{l,\cdot}V^\H D_{ S_{l,\cdot}} D_{ A_{i,\cdot}   } Q_{-l} D_{B_{\cdot,i} } D_{S_{l,\cdot}}VL_{l,\cdot}^\H Q_{ij}\right] S_{l,i}\right)^2 \\
    &\leq \frac1{p^2} \sum_{l'=1}^n \|L_{l',\cdot}\|^2 \frac1{p^2} \sum_{l=1}^p \EE[|Q_{ij}|^2 L_{l,\cdot}D_{S_{\cdot,i}}V^\H D_{S_{l,\cdot}}D_{B_{\cdot,i} } \nonumber \\
    &Q_{-l}^\H D_{ S_{l,\cdot} } D_{ A_{i,\cdot}   }VV^\H D_{ S_{l,\cdot} } D_{ A_{i,\cdot}   } Q_{-l} D_{B_{\cdot,i} }D_{S_{l,\cdot}}VD_{S_{\cdot,i}}L_{l,\cdot}^\H] \\
    &\leq\tr(L^\H L) \frac1{p^4} \sum_{l=1}^p \left\| \EE[ |Q_{ij}|^2 D_{S_{\cdot,i}}V^\H D_{S_{l,\cdot}}D_{B_{\cdot,i} }Q_{-l}^\H D_{ S_{l,\cdot} } D_{ A_{i,\cdot}   }VV^\H D_{ S_{l,\cdot} } D_{ A_{i,\cdot}   } Q_{-l}\right. \nonumber \\
    &\left. D_{B_{\cdot,i} } D_{S_{l,\cdot}}VD_{S_{\cdot,i}} ] \right\| \|L_{l,\cdot}\|^2 \\
    &\leq \frac{C}{p^2} (\tr(L^\H L))^2 = O(p^{-2})
\end{align*}
for $C>0$ a bound on the norm of the matrix in the expectation term. This bound holds because we imposed that $L^\H L\to \mathcal L=O_{\|\cdot\|}(1)$, because $\|Q_{-l}\|\leq 1/|\Im[z]|$ (or $\leq 1/z$ for $z<0$) and because all entries of $S$, $A$, $B$ are bounded.

Therefore, the term in the first line parentheses above is of order $O(p^{-1})$. As a consequence,
\begin{align*}
    &\frac1{p^2} \sum_{l=1}^p\EE \left[ X_{l,\cdot} D_{ S_{l,\cdot} } D_{ A_{i,\cdot}   } Q_{-l} D_{B_{\cdot,i} } D_{S_{l,\cdot}} X_{l,\cdot}^\H Q_{ij}\right] S_{l,i} \\
    &=\frac1{p^2} \sum_{l=1}^p\EE \left[\tr D_{ R_{l,\cdot} } D_{A_{i,\cdot} } Q_{-l} D_{B_{\cdot,i} } D_{S_{l,\cdot}} Q_{ij}\right] S_{l,i} + O(p^{-1}).
\end{align*}

In the remainder of the derivations, we will often use the Cauchy-Schwarz and norm inequalities for more complex terms. We will not further develop them in detail when the result is immediate or close to the previous derivation.

\medskip

Back to our original sum over $(l,l)$ indices, we are now left to estimating the newly introduced quantity
\begin{align*}
    &-\frac1{p^2} X_{l,\cdot} D_{ R_{l,\cdot} } D_{ A_{i,\cdot}   }
    Q_{-l} \left\{\left[D_{S_{l,\cdot}} X_{l,\cdot}'X_{l,\cdot}D_{S_{l,\cdot}}\right]\odot B \right\}\nonumber \\
    &\times\left(I_n + \frac1pQ_{-l} \left\{\left[D_{S_{l,\cdot}} X_{l,\cdot}'X_{l,\cdot}D_{S_{l,\cdot}}\right]\odot B\right\} \right)^{-1} Q_{-l} D_{B_{\cdot,i} } D_{S_{l,\cdot}} X_{l,\cdot}'.
\end{align*}

This term is delicate as the dependence of the inner-matrix in $X_{l,\cdot}$ remains. Here the main observation to make is the following and depends on the nature of $B$:
\begin{align*}
    \frac1p\left\{\left[D_{S_{l,\cdot}} X_{l,\cdot}^\H X_{l,\cdot}D_{S_{l,\cdot}}\right]\odot B \right\} &= \frac1p\left\{\left[D_{S_{l,\cdot}} X_{l,\cdot}^\H X_{l,\cdot}D_{S_{l,\cdot}}\right]\odot \varepsilon_B 1_n1_n^\T \right\} \\
    &+ \frac1p\left\{\left[D_{S_{l,\cdot}} X_{l,\cdot}^\H X_{l,\cdot}D_{S_{l,\cdot}}\right]\odot \mathring{B} \right\} \\
    &+ \frac1p\left\{\left[D_{S_{l,\cdot}} X_{l,\cdot}^\H X_{l,\cdot}D_{S_{l,\cdot}}\right]\odot (b-\varepsilon_B)I_p \right\}
\end{align*}
where we wrote $\mathring{B}=B-\EE[B]$ and used $\EE[B]=\varepsilon_B 1_n1_n^\T+(b-\varepsilon_B)I_p$.

Remark that
\begin{align*}
    \frac1p\left[D_{S_{l,\cdot}} X_{l,\cdot}^\H X_{l,\cdot}D_{S_{l,\cdot}}\right]\odot \mathring{B} &= D_{D_{S_{l,\cdot}}X_{l,\cdot}^\H} \mathring{B}D_{X_{l,\cdot}D_{S_{l,\cdot}}}
\end{align*}
the spectral norm of which is bounded, for all large $n,p$ with high probability by $O(\log p/\sqrt p)$: this is because the spectrum of $\mathring{B}$ follows a semi-circle distribution in the limit with $\|\mathring{B}\|/\sqrt{2n}\to 1$, and $\|D_{X_{l,\cdot}}\|$ is the maximum of $n$ independent Gaussian variables which, uniformly on $X$ cannot grow faster than $O(\sqrt{\log(np)})=O(\sqrt{\log p})$. This claim is confirmed by a further application of the Nash-Poincar\'e inequality. Similarly, $\frac1p\{[D_{S_{l,\cdot}} X_{l,\cdot}^\H X_{l,\cdot}D_{S_{l,\cdot}}]\odot (b-\varepsilon_B)I_p \}$ is bounded in norm by $O(\log p/p)$.

With these remarks at hand, we may freely replace $B$ in the expression of $[D_{S_{l,\cdot}} X_{l,\cdot}^\H X_{l,\cdot}D_{S_{l,\cdot}}]\odot B $ above by $\varepsilon_B 1_n1_n^\T$, so to obtain
\begin{align*}
    &-\frac1{p^2} X_{l,\cdot} D_{ S_{l,\cdot} } D_{ A_{i,\cdot}   }
    Q_{-l} \left\{\left[D_{S_{l,\cdot}} X_{l,\cdot}^\H X_{l,\cdot}D_{S_{l,\cdot}}\right]\odot B \right\}\nonumber \\
    &\times\left(I_n + \frac1pQ_{-l} \left\{\left[D_{S_{l,\cdot}} X_{l,\cdot}^\H X_{l,\cdot}D_{S_{l,\cdot}}\right]\odot B\right\} \right)^{-1}Q_{-l} D_{B_{\cdot,i} } D_{S_{l,\cdot}} X_{l,\cdot}^\H \\
    &=-\frac{\varepsilon_B}{p^2} X_{l,\cdot} D_{ R_{l,\cdot} } D_{ A_{i,\cdot}   }
    Q_{-l} D_{S_{l,\cdot}} X_{l,\cdot}^\H X_{l,\cdot}D_{S_{l,\cdot}} \nonumber \\
    &\times\left(I_n + \frac{\varepsilon_B}pQ_{-l} D_{S_{l,\cdot}} X_{l,\cdot}^\H X_{l,\cdot}D_{S_{l,\cdot}}\right)^{-1}Q_{-l} D_{B_{\cdot,i} } D_{S_{l,\cdot}} X_{l,\cdot}^\H+O_p(\sqrt{\log p}/\sqrt p).
\end{align*}

Using Sherman-Morrison's identity $u^\H(A+\lambda uv^\H)^{-1}=\frac{u^\H A^{-1}}{1+\lambda v^\H A^{-1}u}$, this further simplifies into
\begin{align*}
    &-\frac1{p^2} X_{l,\cdot} D_{ S_{l,\cdot}}D_{ A_{i,\cdot}   }
    Q_{-l} \left\{\left[D_{S_{l,\cdot}} X_{l,\cdot}^\H X_{l,\cdot}D_{S_{l,\cdot}}\right]\odot B \right\}\nonumber \\
    &\times \left(I_n + \frac1pQ_{-l} \left\{\left[D_{S_{l,\cdot}}X_{l,\cdot}^\H X_{l,\cdot}D_{S_{l,\cdot}}\right]\odot B\right\} \right)^{-1}Q_{-l} D_{B_{\cdot,i} } D_{S_{l,\cdot}} X_{l,\cdot}^\H \\
    &=-\frac{\varepsilon_B}{p^2} X_{l,\cdot} D_{ S_{l,\cdot} } D_{ A_{i,\cdot}   }
    Q_{-l} D_{S_{l,\cdot}} X_{l,\cdot}^\H\frac{X_{l,\cdot}D_{S_{l,\cdot}}Q_{-l} D_{B_{\cdot,i} } D_{S_{l,\cdot}} X_{l,\cdot}^\H}{1+\frac{\varepsilon_B}p X_{l,\cdot}D_{S_{l,\cdot}}Q_{-l}D_{S_{l,\cdot}}X_{l,\cdot}^\H} +O_p(\sqrt{\log p}/\sqrt p).
\end{align*}

The quadratic forms are now all accessible and all converge to their traces at uniform speed $O(\log(p)/\sqrt{p})$ (again by a control of their variances), so that
\begin{align*}
    &-\frac1{p^2} X_{l,\cdot} D_{ S_{l,\cdot} } D_{ A_{i,\cdot}   }
    Q_{-l} \left\{\left[D_{S_{l,\cdot}} X_{l,\cdot}^\H X_{l,\cdot}D_{S_{l,\cdot}}\right]\odot B \right\}\nonumber \\
    &\times\left(I_n + \frac1pQ_{-l} \left\{\left[D_{S_{l,\cdot}}X_{l,\cdot}^\H X_{l,\cdot}D_{S_{l,\cdot}}\right]\odot B\right\} \right)^{-1} Q_{-l} D_{B_{\cdot,i} } D_{S_{l,\cdot}} X_{l,\cdot}^\H \\
    &=-\frac{\varepsilon_B}{p^2} \tr D_{S_{l,\cdot}} D_{ A_{i,\cdot} } Q_{-l} \frac{ \tr D_{B_{\cdot,i} } D_{S_{l,\cdot}}^2 Q_{-l} }{1+\frac{\varepsilon_B}p \tr D_{S_{l,\cdot}}Q_{-l}} +O_p(\sqrt{\log p}/\sqrt p).
\end{align*}

With the same argument as above, one may freely replace $Q_{-l}$ by $Q$ up to a negligible cost of $O(1/\sqrt p)$ in the above traces. Then, perturbing matrix $K$ so to discard the contribution of $S_{l,\cdot}$ and $B_{\cdot,i}$ also comes at a negligible cost, so that, again with the same perturbation argument, we get
\begin{align*}
    &-\frac1{p^2} X_{l,\cdot} D_{ S_{l,\cdot} } D_{ A_{i,\cdot} } Q_{-l} \left\{\left[D_{S_{l,\cdot}} X_{l,\cdot}^\H X_{l,\cdot}D_{S_{l,\cdot}}\right]\odot B \right\} \nonumber \\
    &\times\left(I_n + \frac1pQ_{-l} \left\{\left[D_{S_{l,\cdot}} X_{l,\cdot}^\H X_{l,\cdot}D_{S_{l,\cdot}}\right]\odot B\right\} \right)^{-1} Q_{-l} D_{B_{\cdot,i} } D_{S_{l,\cdot}} X_{l,\cdot}^\H \\
    &=-\frac{\varepsilon_B}{p^2} \tr D_{ S_{l,\cdot} } D_{ A_{i,\cdot} } Q_{-l} \frac{ \varepsilon_B\varepsilon_S \tr Q }{1+\frac{\varepsilon_B\varepsilon_S}p \tr Q} +O_p(\sqrt{\log p}/\sqrt p).
\end{align*}

Summarizing the results above, we then get (with the same necessary controls by the Nash-Poincar\'e inequality as above),
\begin{align*}
    T_1(A,R)+T_2(A,R) &= \EE\left[ \frac1p[S^\H S]_{ii}A_{ii}Q_{ij} \right] - \EE\left[ \frac1{p^2}\sum_{l=1}^p \tr \left(D_{S_{l\cdot}} D_{ A_{i\cdot}} Q_{-l} D_{B_{\cdot i}}D_{S_{l\cdot}}\right) S_{li} Q_{ij}\right] \\
    &+ \frac1p\sum_{l=1}^p \EE\left[ \frac{\varepsilon_B}{p^2} \tr \left(D_{S_{l,\cdot}} D_{A_{i,\cdot}}Q_{-l}\right) \frac{ \varepsilon_B\varepsilon_S \tr Q }{1+\frac{\varepsilon_B\varepsilon_S}p \tr Q} S_{li} Q_{ij}\right] + O\left(\frac{\log p}{\sqrt{p}}\right).
\end{align*}

\subsection*{Term 3: $(P^\H,Z)$}

We consider now $T_3(A,S) \equiv \EE\left[ \left[\left(\left[\frac1p(P\odot S)^\H(Z\odot S)\right]\odot A\right) Q\right]_{ij} \right]$:
\begin{align*}
     T_3(A,R) &= \frac1p \sum_{l=1}^p \sum_{m=1}^n P_{li} S_{li} S_{lm} A_{im} \EE \left[ \frac{\partial Q_{mj}}{\partial \bar Z_{lm}}\right]
\end{align*}
with
\begin{align*}
    \EE \left[ \frac{\partial Q_{cd}}{\partial \bar Z_{ab}}\right] &= - \frac{1}{p} \left[Q \left[(E_{ba} \odot S^\H) (X\odot S) \cdot B\right]  Q\right]_{cd}\\
    &= - \frac{1}{p} \sum_{l=1}^n \sum_{m=1}^p Q_{cl} \left[(E_{ba} \odot S^\H) (X\odot S) \cdot B\right]_{lm} Q_{md}\\
    &= - \frac{1}{p} \sum_{l=1}^n \sum_{m=1}^p \sum_{o=1} Q_{cl} \delta_{bl} \delta_{ao} S^\H_{lo}  [X\odot S]_{om}  B_{lm} Q_{md}\\
    &= - \frac{1}{p} \sum_{l=1}^n \sum_{m=1}^p Q_{cl} \delta_{bl} S^\H_{la}  [X\odot S]_{am}  B_{lm} Q_{md}\\
    &= - \frac{1}{p}  \sum_{m=1}^p Q_{cb} S^\H_{ba}  [X\odot S]_{am}  B_{bm} Q_{md}\\
    &= - \frac{1}{p} Q_{cb} S^\H_{ba}  \left[ (X\odot S) D_{B_{b\cdot}} Q\right]_{ad}.
\end{align*}
As a consequence,
\begin{align*}
     T_3(A,R) &= -\frac{1}{p^2} \sum_{l=1}^p \sum_{m=1}^n
     P_{li} S_{li} S_{lm} A_{im} Q_{mm} S_{ml}  \EE\left[ (X\odot S) D_{B_{m\cdot}} Q\right]_{lj}\\
     &= -\frac{1}{p^2} \sum_{l=1}^p P_{li} S_{li} \EE\left[(X\odot S)
     \left(\sum_{m=1}^p  D_{B_{m\cdot}} S_{lm} A_{im} Q_{mm} S_{ml} \right)
     Q \right]_{lj} \\
     &=-\frac{1}{p^2} \sum_{l=1}^p P_{li} S_{li} \EE\left[(X\odot S)
     D_{ \{ \tr Q D_{B_{\cdot s}} D_{S_{l\cdot}} D_{A_{i\cdot}} \}_{s=1}^n } Q \right]_{lj} \\
     &= -\frac{1}{p^2} \sum_{l=1}^p
     \EE\left[ [(T \odot S)^\H]_{il}  \left[(X\odot S)
     D_{ \{ \tr Q D_{B_{\cdot s}} D_{S_{l\cdot}} D_{A_{i\cdot}} \}_{s=1}^n }
     Q \right]_{lj} \right].
\end{align*}

At this stage in the calculus, it is necessary to study the normalized trace
\begin{align*}
    \frac1p\tr Q D_{B_{\cdot s}} D_{A_{i\cdot}} D_{S_{l\cdot}}.
\end{align*}
Note that, unless $B_{\cdot s}^\H=A_{i\cdot}$ (which could only occur for one value of $s$; typically $s=i$ if we take $A=B$), using similar perturbation arguments in the large $n,p$ regime as above (the impact of column $B_{\cdot s}$ is negligible in $Q$), we obtain
\begin{align*}
    \frac1p\tr Q D_{B_{\cdot s}} D_{A_{i\cdot}} D_{S_{l\cdot}} &= \frac{\varepsilon_B}p\tr Q D_{A_{i\cdot}} D_{S_{l\cdot}} + o_p(1).
\end{align*}

As such, $D_{ \{ \tr Q D_{B_{\cdot s}}D_{A_{i\cdot}}D_{S_{l\cdot}} \}_{s=1}^p }$ is asymptotically close to a scaled identity matrix \emph{depending on $i$} and we may then rewrite
\begin{align*}
    T_3(A,R) &= -\frac{\varepsilon_B}{p}\EE\left[ (P \odot R)^\H D_{\{\frac1p\tr Q D_{A_{i\cdot}}D_{S_{l\cdot}}\}_{l=1}^p} (X\odot S) Q \right]_{ij} + o(1).
\end{align*}
This further boils down to
\begin{align*}
    T_3(A,S) &= -\frac{\varepsilon_B\varepsilon_S}{p}\EE\left[D_{\{\frac1p\tr D_{A_{i\cdot}}Q\}_{i=1}^n} (P \odot S)^\H(X\odot S) Q \right]_{ij} + o(1) \\
    &=-\frac{\varepsilon_B\varepsilon_S}{p}\EE\left[\left\{\left[ (P \odot S)^\H(X\odot S) \right] \odot d_{\{\frac1p\tr D_{A_{i\cdot}}Q\}_{i=1}^n} 1_n^\T \right\} Q \right]_{ij} + o(1)
\end{align*}
where $d_v$ is the (column) vector composed of the elements $v_i$.

\subsection*{Term 4: $(P^\H,P)$}

We finally add up the easiest term
\begin{align*}
     T_4(A,R) &\equiv \EE\left[ \left[\left(\left[\frac1p(P\odot S)^\H(P\odot S)\right]\odot A\right) Q\right]_{ij} \right].
\end{align*}

\subsection*{Collecting the terms}

Collecting all terms $T_i(A,S)$, it appears that the desired evaluation of $\EE[Q_{ij}]$, which we obtain through that of
\begin{align*}
    \EE\left[ \frac1p (X \odot S)^\H(X\odot S) Q \right]_{ij}
\end{align*}
gives rise to two sets of ``new'' terms:
\begin{enumerate}
    \item the traces
    \begin{align*}
        \tr \left(D_{S_{l\cdot}}D_{A_{i\cdot}}Q_{-l}\right)
    \end{align*}
    \item the matrix expectation
    \begin{align*}
        \EE\left[\left\{\left[ (P \odot S)^\H(X\odot S) \right] \odot d_{\{\frac1p\tr D_{A_{i\cdot}}Q\}_{i=1}^n} 1_n^\T \right\} Q \right]_{ij}.
    \end{align*}
\end{enumerate}
Using perturbation arguments, the traces are easily analyzed and all lead to scaled versions of $\tr Q$ in the limit, thereby effectively not providing any new term.

The matrix expectation is less immediate and must be appropriately used to ``close the loop'' of the estimate of $\EE[Q_{ij}]$. Specifically,
\begin{itemize}
    \item letting $A=B$ in the initial equation leads to evaluating
    \begin{align*}
        \EE\left[\left\{\left[ (P \odot S)^\H(X\odot S) \right] \odot d_{\{\frac1p\tr D_{B_{i\cdot}}Q\}_{i=1}^n} 1_n^\T \right\} Q \right]_{ij}
\end{align*}
which, from the fact that $\frac1p\tr D_{B_{i\cdot}}Q = \frac{\varepsilon_B}p\tr Q + o_p(1)$, is essentially
\begin{align*}
    \EE\left[ \frac{\varepsilon_B}p\tr Q \left\{\left[ (P \odot S)^\H(X\odot S) \right] \odot 1_n 1_n^\T \right\} Q \right]_{ij}
\end{align*}
a term that we thus need to evaluate;
    \item letting then $A=1_n1_n^\T$ leads instead to
\begin{align*}
    &\EE\left[\left\{\left[ (P \odot S)^\H(X\odot S) \right] \odot d_{\{\frac1p\tr D_{1_n}Q\}_{i=1}^n} 1_n^\T \right\} Q \right]_{ij} \\ &= \EE\left[ ((1/p)\tr Q) \left\{\left[ (P \odot S)^\H(X\odot S) \right] \odot 1_n1_n^\T \right\} Q \right]_{ij}
\end{align*}
from which we may now close the loop.
\end{itemize}

\medskip

Precisely, combining terms from $T_3(1_n1_n^\T,S)$ and $T_4(1_n1_n^\T,S)$, we first find that
\begin{align*}
    &\left\{ \left[ \frac1p (P\odot S)^\H(X\odot S)\right] \odot 1_n1_n^\T \right\} Q \\ \leftrightarrow &-\left(\frac{\varepsilon_B\varepsilon_S}p\tr Q \right) \left\{ \left[ \frac1p (T\odot S)^\H(X\odot S)\right] \odot 1_n1_n^\T \right\} Q\nonumber \\
    &+ \left\{ \left[ \frac1p (P\odot S)^\H(P\odot S)\right] \odot 1_n1_n^\T \right\} Q
\end{align*}
so that, letting $m(z)\in\mathbb C$ be such that $\frac1n\tr Q\leftrightarrow m(z)$,
\begin{align*}
    &\left\{ \left[ \frac1p (P\odot S)^\H(X\odot S)\right] \odot 1_n1_n^\T \right\} Q \nonumber \\
    &\leftrightarrow \frac1{1+c_0^{-1}\varepsilon_B\varepsilon_S m(z)} \left\{ \left[ \frac1p (P\odot S)^\H(P\odot S)\right] \odot 1_n1_n^\T \right\} Q.
\end{align*}

Next, combining all $T_i(B,S)$, we get, using in particular $\frac1p[S^\H S]_{ii}\to \varepsilon_S$ almost surely,
\begin{align*}
    Q &\leftrightarrow -\frac1zI_n + \frac{\varepsilon_Sb}z Q - \varepsilon_S^2\varepsilon_B c_0^{-1}m(z) Q + \frac{\varepsilon_B^3\varepsilon_S^3c_0^{-2}m(z)^2}{1+c_0^{-1}\varepsilon_B\varepsilon_Sm(z)} Q \\
    &- \frac{\varepsilon_B^2\varepsilon_Sc_0^{-1}m(z)}{1+c_0^{-1}\varepsilon_B\varepsilon_Sm(z)} \frac1p (P\odot S)^\H(P\odot S)Q + \frac1p \left\{ \left[(P\odot S)^\H(P\odot S)\right]\odot B \right\}Q.
\end{align*}

\medskip

To complete the proof, we now need to handle the terms $(P\odot S)^\H(P\odot S)Q$ and $\frac1p\{ [(P\odot S)^\H(P\odot S)]\odot B\}Q$ and relate then to $Q$ directly. To this end, similar to previously, let us write $S=\varepsilon_S 1_p1_n^\T+\mathring{S}$, $B=\varepsilon_B1_n1_n^\T+\mathring{B}$ and $P=\sum_{\ell=1}^kL_{\cdot,\ell} V_{\cdot,\ell}^\H$. Then, since we imposed the entries $V_{ij}$ to be essentially of order $1/\sqrt n$,\footnote{More specifically, it is enough to assume that $V_{ij}^2=o(1/\sqrt n)$.} observe that the matrix $\frac1{\sqrt p} L_{\cdot,\ell} V_{\cdot,\ell}^\H \odot \mathring{S}=\frac1{\sqrt p} D_{L_{\cdot,\ell}} \mathring{S} D_{V_{\cdot,\ell}^\H}$ has operator norm of order $1/\sqrt p$. The same reasoning applies to $B$, so that we may rewrite
\begin{align*}
    zQ \leftrightarrow & -I_n + \varepsilon_Sb Q - \varepsilon_S^2\varepsilon_B c_0^{-1}m(z) Q + \frac{\varepsilon_B^3\varepsilon_S^3c_0^{-2}m(z)^2}{1+c_0^{-1}\varepsilon_B\varepsilon_Sm(z)} Q \\
    &- \frac{\varepsilon_B^2\varepsilon_S^3c_0^{-1}m(z)}{1+c_0^{-1}\varepsilon_B\varepsilon_Sm(z)} \frac1p P^\H PQ + \varepsilon_B\varepsilon_S^2\frac1p P^\H PQ \\
    =&-I_n + \varepsilon_Sb Q - \varepsilon_S^2\varepsilon_B c_0^{-1}m(z) Q + \frac{\varepsilon_B^3\varepsilon_S^3c_0^{-2}m(z)^2}{1+c_0^{-1}\varepsilon_B\varepsilon_Sm(z)} Q \\
    &+ \frac{\varepsilon_B\varepsilon_S^2}{1+c_0^{-1}\varepsilon_B\varepsilon_Sm(z)} \frac1p P^\H PQ
\end{align*}


Further using
\begin{align*}
    \frac1pP^\H P &= \frac1p V L^\H L V^\H = V \mathcal L V^\H + o_{\|\cdot\|}(1)
\end{align*}
along with the fact that $V^\H V=I_k$, we finally get the deterministic equivalent for $Q$
\begin{align*}
    Q \leftrightarrow &\left[(\varepsilon_Sb-z)I_n-c_0^{-1}\varepsilon_S^2\varepsilon_Bm(z) I_n + \frac{c_0^{-2}\varepsilon_S^3\varepsilon_B^3m(z)^2}{1+\varepsilon_B\varepsilon_Sc_0^{-1}m(z)}I_n \right. \nonumber \\
    &\left. + \frac{c_0^{-1}\varepsilon_S^2\varepsilon_B}{1+\varepsilon_B\varepsilon_Sc_0^{-1}m(z)}V\mathcal LV^\H\right]^{-1}.
\end{align*}

In particular, recalling that $m(z)$ is an asymptotic equivalent for $\frac1n\tr Q$, we have
\begin{align*}
    m(z) &= \frac1{(\varepsilon_Sb-z)-c_0^{-1}\varepsilon_S^2\varepsilon_B m(z)+ \frac{c_0^{-2}\varepsilon_S^3\varepsilon_B^3m(z)^2}{1+\varepsilon_B\varepsilon_Sc_0^{-1}m(z)}}
\end{align*}
which unfolds from $V\mathcal{L}V^\H$ being of finite rank $k$ (so that it does not affect the limiting normalized trace) and thus provides a \emph{deterministic equivalent}. Equivalently, this is
\begin{align*}
    z = \varepsilon_Sb - \frac1{m(z)} - c_0^{-1}\varepsilon_B\varepsilon_S^2m(z)+\frac{c_0^{-2}\varepsilon_B^3\varepsilon_S^3 m(z)^2}{1+c_0^{-1}\varepsilon_B\varepsilon_Sm(z)}
\end{align*}
which, integrated in the previous expression of the random equivalent of $Q$, provides the shorter and final forms of the \emph{deterministic equivalent}:
\begin{align*}
    \boxed{
    Q \leftrightarrow  m(z) \left[I_n + \frac{c_0^{-1}\varepsilon_S^2\varepsilon_B m(z)}{1+\varepsilon_B\varepsilon_Sc_0^{-1}m(z)}V\mathcal LV^\H\right]^{-1}
    }
\end{align*}
or possibly more expressively
\begin{align*}
    Q &\leftrightarrow m(z) V_\perp V_{\perp}^\H + m(z)V\left[ I_k+ \frac{c_0^{-1}\varepsilon_S^2\varepsilon_B m(z)}{1+\varepsilon_B\varepsilon_Sc_0^{-1}m(z)}\mathcal L \right]^{-1}V^\H
\end{align*}
where in this last equality $V_\perp$ is an orthonormal basis completing $V$ (this last result follows from Woodbury's matrix inverse identity).

\subsection{Proof of Theorem~2}

With the previous result available, the proof of Theorem~2 follows from a classical random matrix approach.

Let $\mathcal L=\sum_{i=1}^{\bar k}\ell_i \mathcal V_i\mathcal V_i^\H$ be the spectral decomposition of $\mathcal L$ with $\mathcal V_i\in\mathbb C^{k\times L_i}$ isometric and such that $\Pi_i=\mathcal V_i\mathcal V_i^\H$ is a projector on the eigenspace associated to eigenvalue $\ell_i$ which we assume of multiplicity $L_i$ greater or equal to $1$.

Then, assuming asymptotic separability (that is, the existence of a spike associated to $\ell_i$), such that the resulting associated eigenvalue(s) $\lambda_j,\ldots,\lambda_{j+L_i-1}$ of $K$ converge to $\rho_i$ with associated eigenspace $\hat{\mathcal V}_i$, we have, in the large $n,p$ limit, almost surely (the limit is needed to ensure that $\lambda_j,\ldots,\lambda_{j+L_i-1}$ fall into the contour $\Gamma_{\rho_i}$),
\begin{align*}
    \hat{\mathcal V}_i\hat{\mathcal V}_i^\H &= -\frac1{2\pi\imath}\oint_{\Gamma_{\rho_i}} Q(z)dz \\
    &\leftrightarrow -\frac1{2\pi\imath}\oint_{\Gamma_{\rho_i}} m(z) V \left[ I_k+ \frac{c_0^{-1}\varepsilon_S^2\varepsilon_B m(z)}{1+\varepsilon_B\varepsilon_Sc_0^{-1} m(z)}\mathcal L \right]^{-1} V^\H dz
\end{align*}
for $\Gamma_x$ a positively oriented complex contour surrounding $x$ closely. By residue calculus, we then find that
\begin{align*}
    \hat{\mathcal V}_i\hat{\mathcal V}_i^\H &\leftrightarrow -\lim_{z\in\mathbb C\to \rho_i} (z-\rho_i) m(z) U \left[ I_k+ \frac{c_0^{-1}\varepsilon_S^2\varepsilon_B m(z)}{1+\varepsilon_B\varepsilon_Sc_0^{-1}m(z)}\mathcal L \right]^{-1} V^\H \\
    &\leftrightarrow -\lim_{z\in\mathbb C\to \rho_i} (z-\rho_i) m(z) V \mathcal V_i \left[ 1 + \frac{c_0^{-1}\varepsilon_S^2\varepsilon_B m(z)}{1+\varepsilon_B\varepsilon_Sc_0^{-1}m(z)}\ell_i \right]^{-1} \mathcal V_i^\H V^\H
\end{align*}
where we exploited the fact that the denominator above must vanish as $z\to\rho_i$, thereby in passing \emph{defining} $\rho_i$.

Specifically, we find that $\rho_i$, the limit of the empirical eigenvalues of $K$ associated with $\ell_i$, is solution to
\begin{align*}
    1 + \ell_i \frac{c_0^{-1}\varepsilon_B\varepsilon_S^2m(\rho_i)}{1+c_0^{-1}\varepsilon_B\varepsilon_S m(\rho_i)}=0 &\Leftrightarrow \frac1{m(\rho_i)} = - c_0^{-1}\varepsilon_B\varepsilon_S(1+\varepsilon_S\ell_i).
\end{align*}
In particular, we have the following convenient relation for what follows:
\begin{align*}
    1+c_0^{-1}\varepsilon_B\varepsilon_S m(\rho_i) &=\frac{\varepsilon_S\ell_i}{1+\varepsilon_S\ell_i}.
\end{align*}
Exploiting the relation $z=f(m(z))$ above, applied to $z=\rho_i$, this leads to the explicit value of the isolated ``spike'' $\rho_i$:
\begin{align*}
    \boxed{\rho_i = \varepsilon_Sb+c_0^{-1}\varepsilon_B\varepsilon_S(1+\varepsilon_S\ell_i)+\frac{\varepsilon_S}{1+\varepsilon_S\ell_i}+\frac{\varepsilon_B}{\ell_i(1+\varepsilon_S\ell_i)}.}
\end{align*}

By l'Hospital's rule (or equivalently a first order Taylor expansion of both numerator and denominator in the inverse formula of the residue), we then have
\begin{align*}
    \hat{\mathcal U}_i\hat{\mathcal V}_i^\H &\leftrightarrow -V \mathcal V_i \frac{m(\rho_i)(1+c_0^{-1}\varepsilon_B\varepsilon_Sm(\rho_i))^2}{\ell_i c_0^{-1}\varepsilon_B\varepsilon_S^2m'(\rho_i)} \mathcal V_i^\H V^\H
\end{align*}
where, exploiting the defining equation of $m(z)$, we find after mere algebraic calculus
\begin{align*}
    \frac1{m'(z)} &= \frac1{m(z)^2} - c_0^{-1}\varepsilon_B\varepsilon_S^2 + c_0^{-2}\varepsilon_B^3\varepsilon_S^3 m(z) \frac{2+c_0^{-1}\varepsilon_B\varepsilon_Sm(z)}{(1+c_0^{-1}\varepsilon_B\varepsilon_Sm(z))^2} \\
    &= \frac1{m(z)^2} - c_0^{-1}\varepsilon_B\varepsilon_S^2 + \frac{c_0^{-2}\varepsilon_B^3\varepsilon_S^3 m(z)}{1+c_0^{-1}\varepsilon_B\varepsilon_S m(z)} + \frac{c_0^{-2}\varepsilon_B^3\varepsilon_S^3 m(z)}{(1+c_0^{-1}\varepsilon_B\varepsilon_S m(z))^2}.
\end{align*}

Altogether, we finally find the fully explicit deterministic equivalent
\begin{align*}
\hspace*{-10mm}\boxed{
    \hat{\mathcal V}_i\hat{\mathcal V}_i^\H \leftrightarrow \left( \frac{\varepsilon_S\ell_i}{1+\varepsilon_S\ell_i} - \frac{\varepsilon_S\ell_i}{c_0^{-1}\varepsilon_B(1+\varepsilon_S\ell_i)^3} - \frac1{c_0^{-1}(1+\varepsilon_S\ell_i)^3} - \frac1{c_0^{-1}\varepsilon_S\ell_i(1+\varepsilon_S\ell_i)^2} \right) V \mathcal V_i\mathcal V_i^\H V^\H.
    }
\end{align*}

Equating the term in parentheses to zero then provides the phase transition condition: indeed, the asymptotic alignment of population and sample eigenspaces vanishes right at the position where the spike $\rho_i$ escapes the limiting continuous part of the support of the eigenvalues of $K$. The phase transition for $\rho_i$ then occurs when $\ell_i$ satisfies:
\begin{align*}
\boxed{
    0=\ell_i^4+\frac{2}{\varepsilon_S}\ell_i^3+\frac{1}{\varepsilon_S^2}\left( 1 - \frac{c_0}{\varepsilon_B}\right)\ell_i^2 - \frac{ 2 c_0}{\varepsilon_S^3}\ell_i-\frac{c_0}{\varepsilon_S^4} \equiv F(\ell_i).
    }
\end{align*}
This expression of $F$ is convenient as it takes the form of a polynomial of order $4$ with unit leading monomial coefficient. It then suffices to remark that the asymptotic alignment expression above expresses as $F(\rho_i)/\ell_i/(1+\varepsilon_S\ell_i)^3$ to conclude the proof of Theorem~2.